\documentclass[10pt,twocolumn,letterpaper]{article}

\usepackage{cvpr}
\usepackage{times}
\usepackage{epsfig}
\usepackage{graphicx}
\usepackage{subfigure}
\usepackage{amsmath}
\usepackage{amssymb}
\usepackage{bm}
\usepackage{authblk}
\usepackage{makecell}
\usepackage{cite}
\usepackage{mathrsfs}
\usepackage{threeparttable}
\usepackage{booktabs}

\usepackage[pagebackref=true,breaklinks=true,letterpaper=true,colorlinks,bookmarks=false]{hyperref}
\cvprfinalcopy 

\def\cvprPaperID{1296} 

\begin{document}

\title{Beyond Tracking: \\ Selecting Memory and Refining Poses for Deep Visual Odometry}

\author[1,3]{Fei Xue} 
\author[1,3]{Xin Wang}
\author[1,3]{Shunkai Li}
\author[1,3]{Qiuyuan Wang} 
\author[2]{Junqiu Wang}
\author[1,3]{Hongbin Zha}
\affil[1]{Key Laboratory of Machine Perception (MOE), School of EECS, Peking University}
\affil[2]{Beijing Changcheng Aviation Measurement and Control Institute, AVIC}
\affil[3] {PKU-SenseTime Machine Vision Joint Lab}
\affil[ ]{\tt\small \{feixue, xinwang\_cis, lishunkai, wangqiuyuan\}@pku.edu.cn \authorcr
 \tt\small jerywangjq@foxmail.com, zha@cis.pku.edu.cn}

\maketitle

\begin{abstract}
Most previous learning-based visual odometry (VO) methods take VO as a pure tracking problem. In contrast, we present a VO framework by incorporating two additional components called Memory and Refining. The Memory component preserves global information by employing an adaptive and efficient selection strategy. The Refining component ameliorates previous results with the contexts stored in the Memory by adopting a spatial-temporal attention mechanism for feature distilling. Experiments on the KITTI and TUM-RGBD benchmark datasets demonstrate that our method outperforms state-of-the-art learning-based methods by a large margin and produces competitive results against classic monocular VO approaches. Especially, our model achieves outstanding performance in challenging scenarios such as texture-less regions and abrupt motions, where classic VO algorithms tend to fail. 
\end{abstract}

\section{Introduction}
Visual Odometry (VO) and Visual Simultaneous Localization And Mapping (V-SLAM) estimate camera poses from image sequences by exploiting the consistency between consecutive frames. 
As an essential task in autonomous driving and robotics, VO has been studied for decades and many outstanding algorithms have been developed \cite{mur2017orb-slam2, engel2014lsd-slam, engel2018dso, wang2017stereo, geiger2011stereoscan}. Recently, as Convolutional Neural Networks (CNNs) and Recurrent Neural Networks (RNNs) achieve impressive performance in many computer vision tasks \cite{yang2018dvso, dosovitskiy2015flownet, brahmbhatt2018mapnet, henriques2018mapnet}, a number of end-to-end models have been proposed for VO estimation. These methods either learn depth and ego-motion jointly with CNNs \cite{zhou2017egomotion, zhan2018feature, yin2018geonet, li2018undeepvo, mahjourian2018vid2depth}, or leverage RNNs to introduce temporal information \cite{wang2017deepvo, wang2018espvo, xue2018fea, parisotto2018gpe, iyer2018ctc}. Due to the high dimensionality of depth maps, the number of frames is commonly limited to no more than 5. Although temporal information is aggregated through recurrent units, RNNs are incapable of remembering previous observations for long time \cite{sukhbaatar2015memory}, leading to the limited usage of historical information. Besides, the above methods pay little attention to the significance of incoming observations for refining previous results, which is crucial for VO tasks. 

\begin{figure}[t]
	\begin{center}
		\includegraphics[width=1.\linewidth]{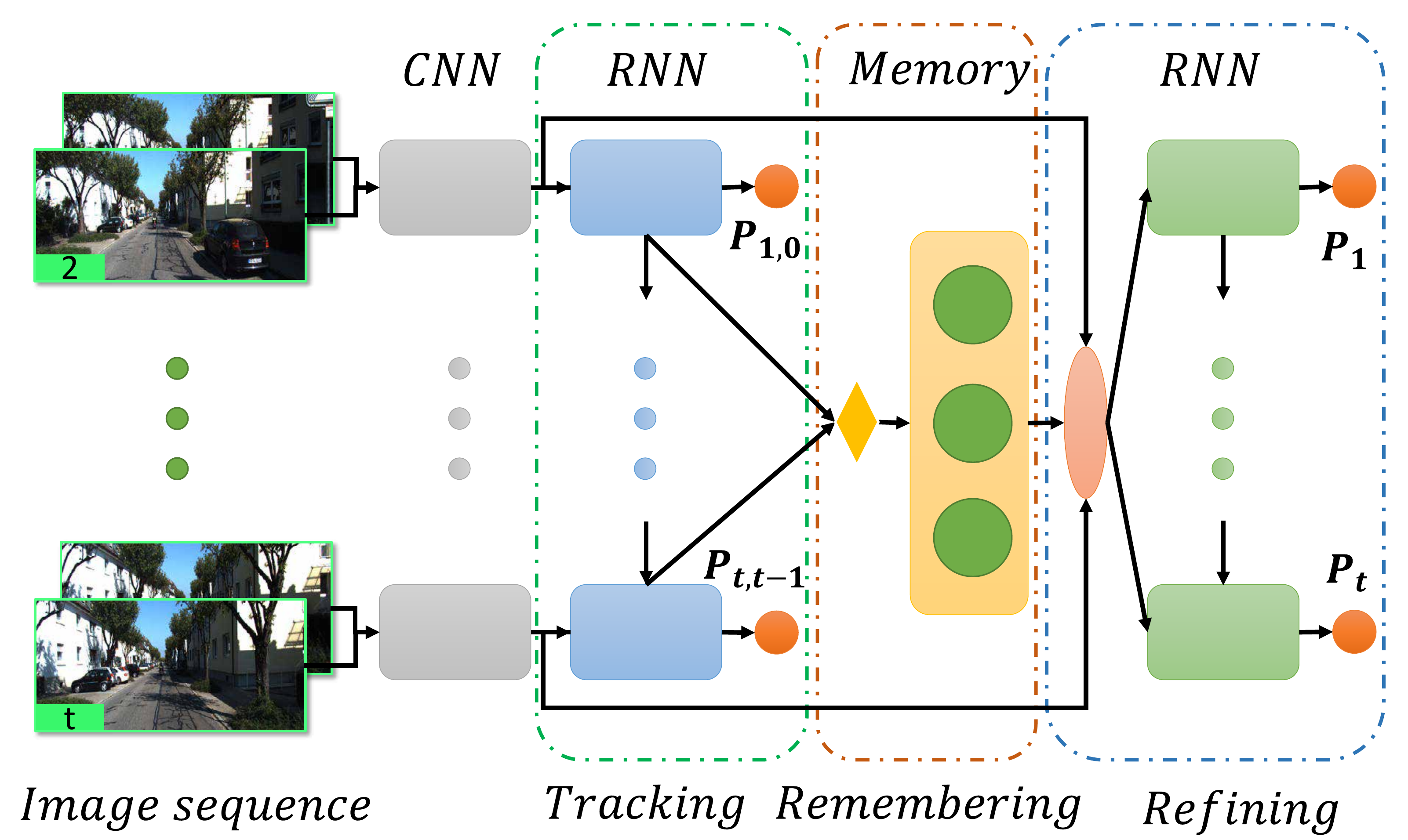}
	\end{center}
	\label{fig:framework}
	\caption{Overview of our framework. Compared with existing learning-based methods which formulate VO task as a pure tracking problem, we introduce two useful components called \textit{Memory} and \textit{Refining}. The \textit{Memory} module preserves longer time information by adopting an adaptive context selection strategy. The \textit{Refining} module ameliorates previous outputs by employing a spatial-temporal feature reorganization mechanism.}
\end{figure}

Direct estimation of camera motions from image snippets is prone to large errors due to the geometric uncertainty caused by \textit{small baselines} (especially for handheld devices). Consequently, error accumulation is getting increasingly severe over time, as global poses are integrated from frame-wise poses. In classic VO/SLAM systems \cite{mur2017orb-slam2}, a local map is established according to the co-visibility graph over up to hundreds of frames, on which bundle adjustment is executed to jointly optimize all corresponding poses. Therefore, both previous and new observations are incorporated for optimization, and accumulated errors are thus alleviated. 

Inspired by classic VO/SLAM systems \cite{mur2017orb-slam2, engel2018dso}, we introduce an effective component, called \textit{Memory}, which explicitly preserves the accumulated information adaptively. Owing to the high sample frequency, contents between consecutive frames are much overlapped. Rather than keeping accumulated information per time step with brute force, an intuitive and efficient strategy is utilized to reduce the redundancy.
As errors of previous poses will propagate over time to current estimation, refining previous results becomes necessary. The \textit{Memory} contains more \textit{global} information, which can be leveraged naturally to refine previous results. Therefore, a \textit{Refining} component is introduced. The \textit{Refining} module takes the global pose estimation as a registration problem by aligning each view with the \textit{Memory}.
A spatial-temporal attention mechanism is applied to the contexts stored in the \textit{Memory} for feature selection.

The overview of our framework is illustrated in Fig.~\ref{fig:framework}. The encoder encodes paired images into high-level features. The \textit{Tracking} module accepts sequential features as input, fuses current observation into accumulated information using convolution LSTMs \cite{xingjian2015convolutional} for preserving spatial connections, and produces relative poses. Hidden states of the \textit{Tracking} RNN are adaptively preserved in the \textit{Memory} slots. The \textit{Refining} module ameliorates previous results using another convolutional LSTM, enabling refined results passing through recurrent units to improve the following outputs. Our contributions can be summarized as follows:
\begin{itemize}
	\item We propose a novel end-to-end VO framework consisting of the \textit{Tracking}, \textit{Memory} and \textit{Refining} components;
	\item An adaptive and efficient strategy is adopted for the \textit{Memory} component to preserve accumulated information;
	\item A spatial-temporal attention mechanism is employed for the \textit{Refining} component to distill valuable features.
\end{itemize}

Our method outperforms state-of-the-art learning-based methods and produces competitive results against classic algorithms. Additionally, it works well in challenging conditions where classic algorithms tend to fail due to insufficient textures or abrupt motions. The rest of this paper is organized as follows. In Sec.~\ref{related_work}, related works on monocular odometry are discussed. In Sec.~\ref{method}, our architecture is described in detail. The performance of the proposed approach is compared with current state-of-the-art methods in Sec.~\ref{experiment}. We conclude the paper in Sec.~\ref{conclusion}.  
\section{Related Works}
\label{related_work}
Visual odometry has been studied for decades, and many excellent approaches have been proposed. Traditionally, VO is tackled by minimizing geometric reprojection errors \cite{mur2017orb-slam2,geiger2011stereoscan, liu2018ice} or photometric errors \cite{engel2014lsd-slam, engel2018dso, wang2017stereo}. These methods mostly work in regular environments, but will fail in challenging scenarios such as textureless scenes or abrupt motions. After the advent of CNNs and RNNs, the VO task has been explored with deep learning techniques. A number of approaches have been proposed to deal with the challenges in classic monocular VO/SLAM systems such as feature detection \cite{agrawal2015learning}, depth initialization \cite{tateno2017cnn-slam, yang2018dvso}, scale correction \cite{yin2017scale}, depth representation \cite{bloesch2018codeslam} and data association \cite{lianos2018vso, bowman2017sslam}. Despite their promising performance, they utilize the classic framework as backend, and thus cannot been deployed in an end-to-end fashion. In this paper, we mainly focus on learning-based end-to-end monocular VO works. \\

\noindent
\textbf{Unsupervised methods} Mimicking the conventional structure from motion, SfmLearner \cite{zhou2017egomotion} learns the single view depth and ego-motion from monocular image snippets using photometric errors as supervisory signals. Following the same scenario, Vid2Depth \cite{mahjourian2018vid2depth} adopts a differential ICP (Iterative Closest Point) loss executed on estimated 3D point clouds to enforce the consistency of predicted depth maps of two consecutive frames. GeoNet \cite{yin2018geonet} estimates the depth, optical flow and ego-motion jointly from monocular views. To cope with the scale ambiguity of motions recovered from monocular image sequences, Depth-VO-Feat \cite{zhan2018feature} and UnDeepVO \cite{li2018undeepvo} extend the work of SfmLearner to accept stereo image pairs as input and recover the absolute scale with the known baseline.

Although these unsupervised methods break the limitation of requiring massive labeled data for training, only a limited number of consecutive frames can be processed in a sequence due to the fragility of photometric losses, leading to high geometric uncertainty and severe error accumulation. \\

\noindent
\textbf{Supervised methods} DeMoN \cite{ummenhofer2017demon} jointly estimates the depth and camera poses in an end-to-end fashion by formulating structure from motion as a supervised learning problem. DeepTAM \cite{zhou2018deeptam} extends DTAM \cite{newcombe2011dtam} via two individual subnetworks indicating \textit{tracking} and \textit{mapping} for the pose and depth estimation respectively. Both DeMoN and DeepTAM achieve promising results, yet require highly labeled data (depth, optical flow and camera pose) for training. MapNet \cite{henriques2018mapnet} presents an allocentric spatial memory for localization, but only discrete directions and positions can be obtained in synthetic environments. 

VO can be formulated as a sequential learning problem via RNNs. DeepVO \cite{wang2017deepvo} harnesses the LSTM \cite{hochreiter1997lstm} to introduce historical knowledge for current relative motion prediction. Based on DeepVO, ESP-VO \cite{wang2018espvo} infers poses and uncertainties in a unified framework. GFS-VO \cite{xue2018fea} considers the discriminability of features to different motion patterns and estimates the rotation and translation separately with a dual-branch LSTM. In addition, the ConvLSTM unit \cite{xingjian2015convolutional} is adopted to retain the spatial connections of features. There are some other works focusing on reducing localization errors by imposing constraints of relative poses \cite{parisotto2018gpe, iyer2018ctc, brahmbhatt2018mapnet}. 

Geometric uncertainty can be partially reduced by aggregating more temporal information using RNNs or LSTMs. Unfortunately, RNNs or LSTMs are limited for remembering long historical knowledge \cite{sukhbaatar2015memory}. Here, we extend the field of view by adaptively preserving hidden states of recurrent units as memories. Therefore, previous valuable information can be inherited longer than being kept in only the single current hidden state. Besides, all these methods ignore the importance of new observations for refining previous poses, which is essential for VO tasks. By incorporating the \textit{Refining} module, previous poses can be updated by aligning filtered features with the \textit{Memory}. Therefore, error accumulation is further mitigated.

\section{Approach}
\label{method}
The encoder extracts high-level features from consecutive RGB images in Sec.~\ref{encoder}. The \textit{Tracking} module accepts sequential features as input, aggregates temporal information, and produces relative poses in Sec.~\ref{tracking}. Hidden states of the \textit{Tracking} RNN are adaptively selected to construct the \textit{Memory} (Sec.~\ref{remembering}) for further \textit{Refining} previous results in Sec.~\ref{refinement}. We design the loss function considering both relative and absolute pose errors in Sec.~\ref{loss_function}.

\subsection{Encoder}
\label{encoder}
We harness CNNs to encode images into high-level features. Optical flow has been proved useful for estimating frame-to-frame ego-motion by lots of current works \cite{wang2017deepvo, wang2018espvo, xue2018fea, zhou2018deeptam, parisotto2018gpe}. We design the encoder based on the Flownet \cite{dosovitskiy2015flownet} which predicts optical flow between two images. The encoder retains the first 9 convolutional layers of Flownet encoding a pair of images, concatenated along RGB channels, into a 1024-channel 2D feature-map. The process can be described as: 
\begin{align}
X_t = \mathcal{F}(I_{t-1}, I_t) \ .
\end{align}
$X_t \in \mathbb{R}^{C \times H \times W} $ denotes the encoded feature-map at time $t$ by function $\mathcal{F}$ from two consecutive images $I_{t-1}$ and $I_t$. $H$, $W$ and $C$ represent the height, width and channel of obtained feature maps.

\begin{figure}[t]
	\begin{center}
		\includegraphics[width=0.95\linewidth]{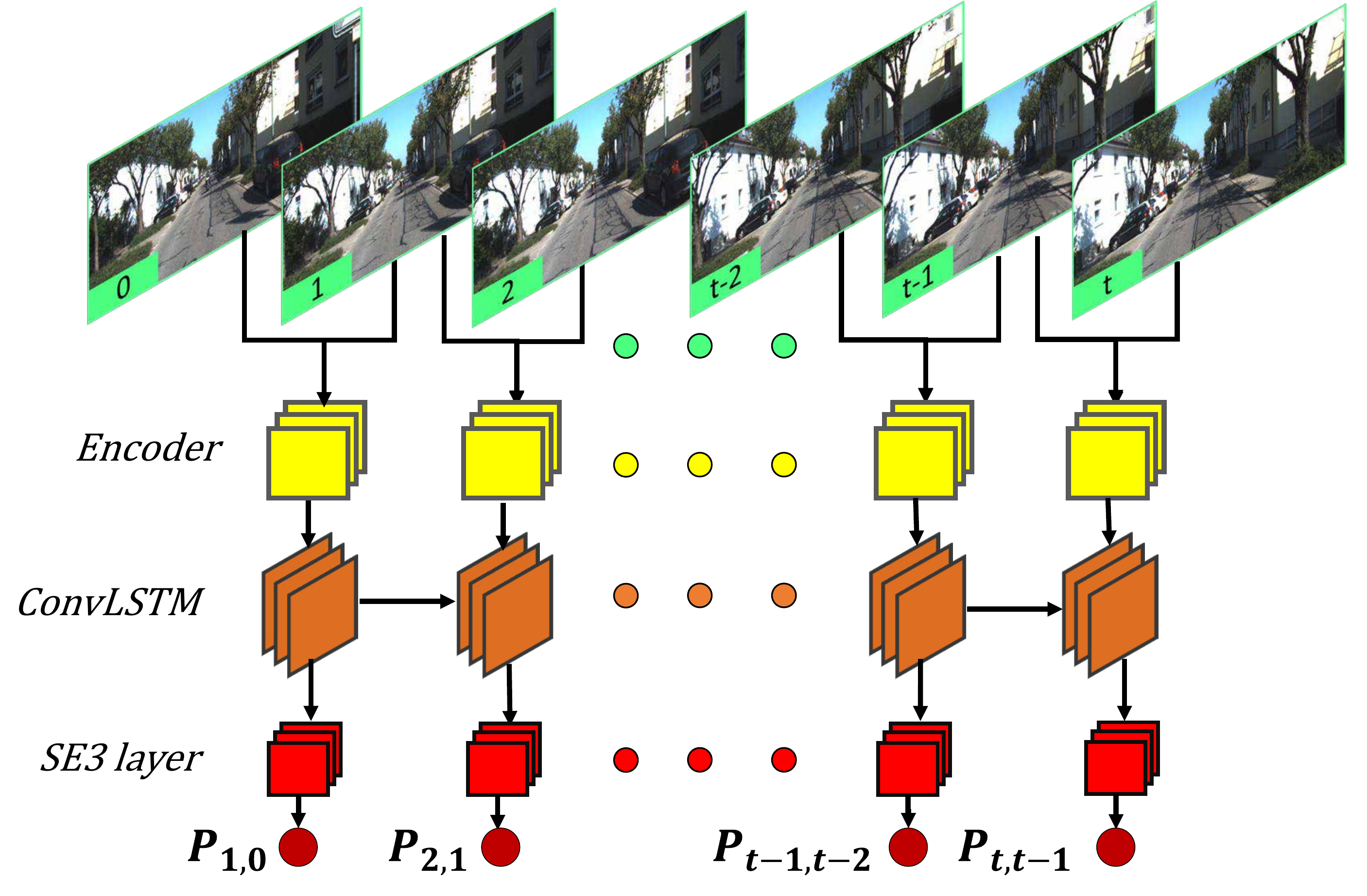}
	\end{center}
	\caption{The \textit{Tracking} module of our framework is implemented on a convolutional LSTM \cite{xingjian2015convolutional}. Relative camera poses are produced by the $\mathbb{SE}$ (3) layer \cite{clark2017vidloc} from the outputs of recurrent units. Temporal information is preserved in the hidden states.}
	\label{fig:tracking}
\end{figure}

\subsection{Tracking}
\label{tracking}
The \textit{Tracking} module fuses current observations into accumulated information and calculates relative camera motions between two consecutive views as shown in Fig.~\ref{fig:tracking}.

\textbf{Sequence modeling} We adopt the  prevent LSTM \cite{hochreiter1997lstm} to model the image sequence. In this case, the feature flow passing through recurrent units carries rich accumulated information of previous inputs to infer the current output. Note that the standard LSTM unit used by DeepVO \cite{wang2017deepvo} and ESP-VO \cite{wang2018espvo} requires 1D vector as input in which the spatial structure of features is ignored. The ConvLSTM unit \cite{xingjian2015convolutional}, an extension of LSTM with convolution underneath, is adopted in the \textit{Tracking} RNN for preserving the spatial formulation of visual cues and expanding the capacity of recurrent units for remembering more knowledge. The recurrent process can be controlled by
\begin{align}
O_t, H_t = \mathcal{U}(X_t, H_{t-1}) \ .
\end{align}
$O_t$ denotes the output at time $t$. $H_{t}$ and $H_{t-1}$ are the hidden states at current and the last time step.

\textbf{Relative pose estimation} Relative motions can be directly recovered from paired images. Unfortunately, direct estimation is prone to error accumulation due to the geometric uncertainty brought by short baselines. The problem can be mitigated by introducing more historical information. Inheriting accumulated knowledge, the output of recurrent unit at each time step is naturally used for pose estimation. The $\mathbb{SE}$ (3) \cite{clark2017vidloc} layer generates the 6-DoF motion $P_{t,t-1}$ from the output at time $t$.

Theoretically, the global pose of each view can be recovered by integrating predicted relative poses as $P_t = \prod_{i=1}^{t}P_{i, i-1}P_0$ ($P_0$ denotes the origin pose of the world coordinate) just as DeepVO \cite{wang2017deepvo} and ESP-VO \cite{wang2018espvo}. The accumulated error, however, will get increasingly severe, and thus degrades the performance of the entire system. Due to the lack of explicit geometric representation of the 3D environments, neural networks, however, are incapable of building a global map to assist tracking. Fortunately, the temporal information is recorded in the hidden states of recurrent units. Although the information is short-time, these hidden states at different time points can be gathered and reorganized as parts of an \textit{implicit} map (discussed in Sec.~\ref{remembering}). 

\begin{figure}[t]
	\centering
	\subfigure[]{
		\begin{minipage}{0.3\textwidth}
			\centering
			\includegraphics[height=4.0cm]{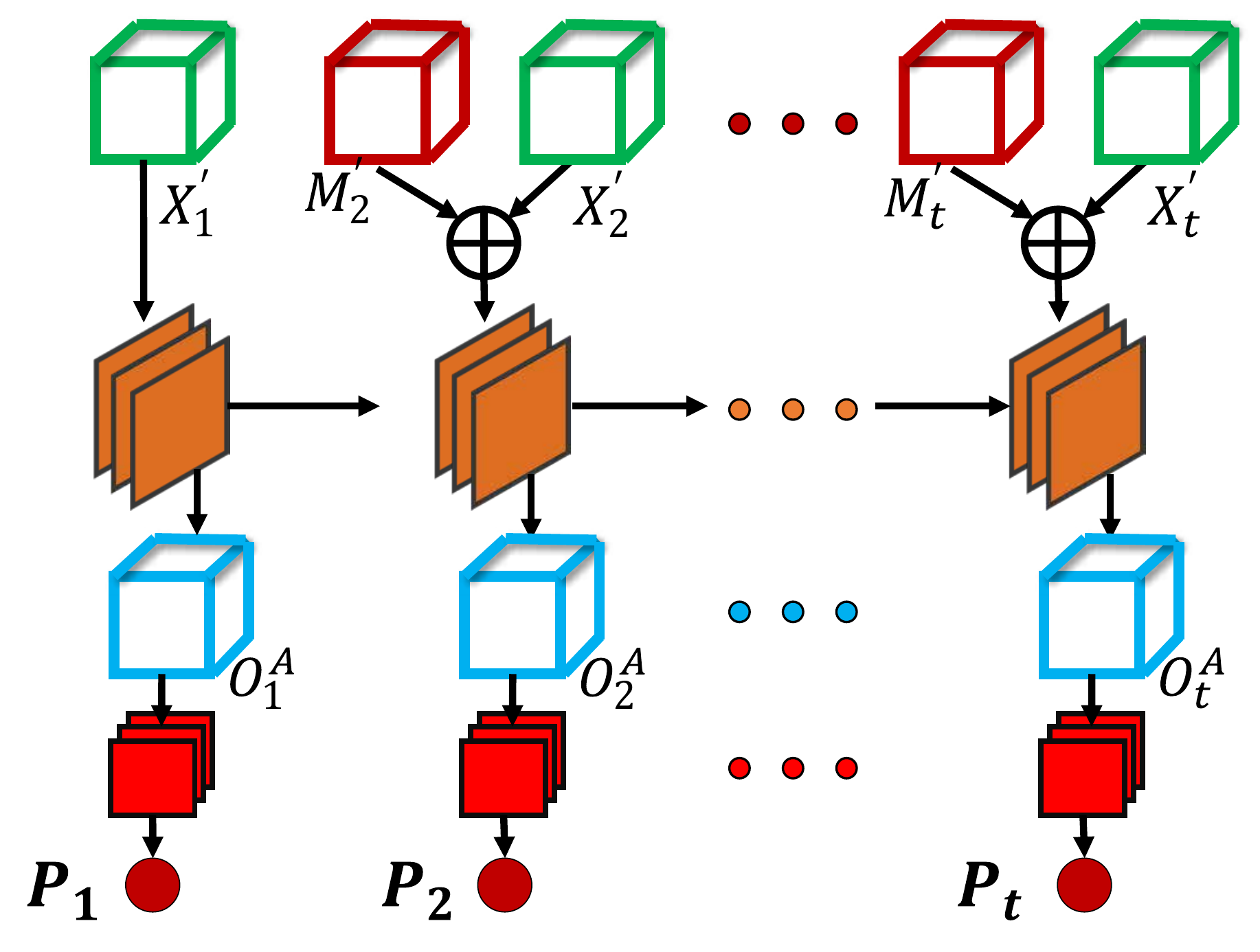}
	\end{minipage}}
	\subfigure[]{
		\begin{minipage}{0.16\textwidth}
			\centering
			\includegraphics[height=4.0cm]{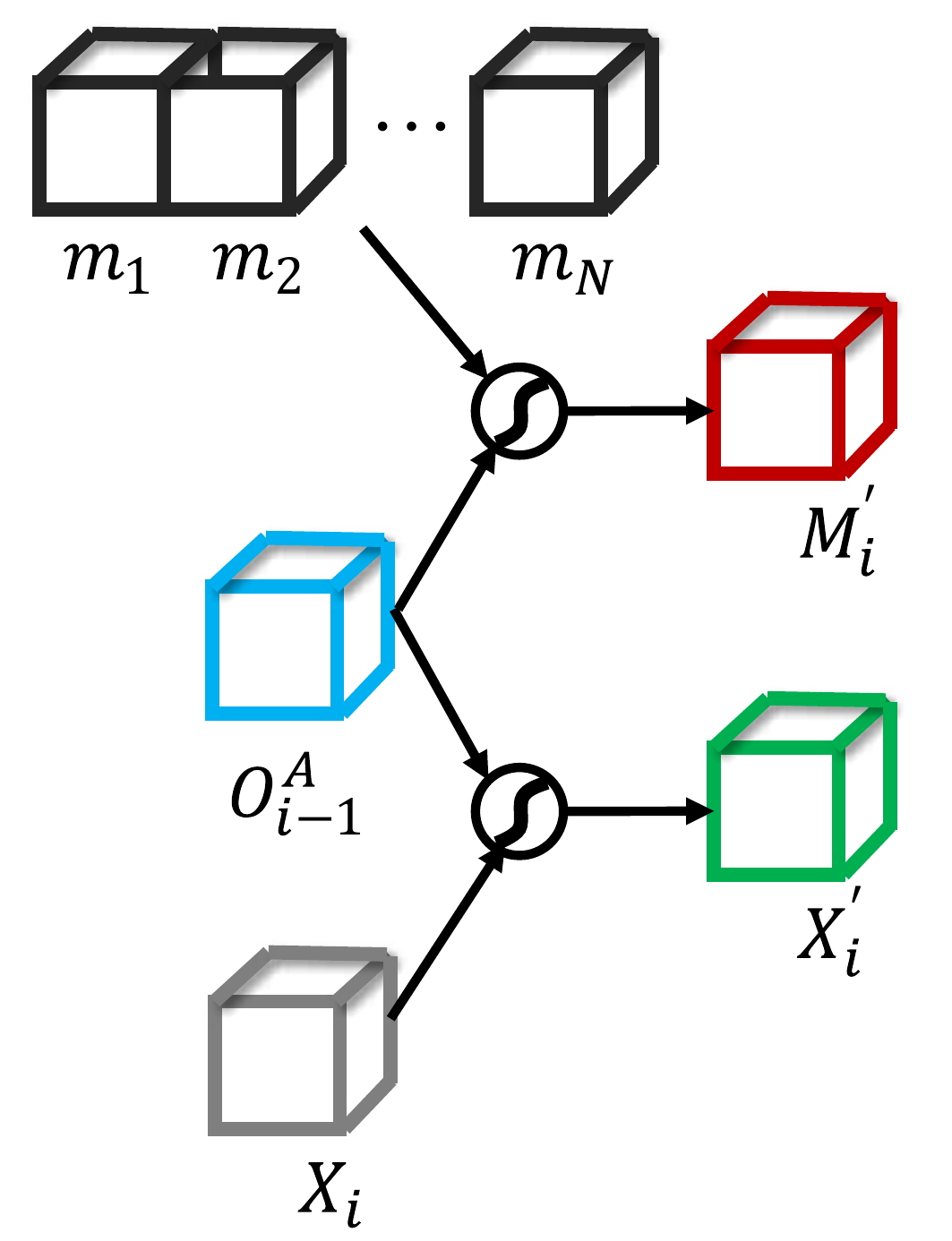}
	\end{minipage}}
	\caption{(a) The \textit{Refining} module aligns current observation with the contexts stored in the \textit{Memory} module for absolute pose estimation. (b) Both contexts and the current observation are re-organized utilizing the last output as guidance.}
	\label{fig:refining}
\end{figure}

\subsection{Remembering}
\label{remembering}
The \textit{Memory} module is a neural analogue of the \textit{local map} commonly used in classic VO/SLAM systems \cite{mur2017orb-slam2}. Considering the LSTM cannot remember information for long time \cite{sukhbaatar2015memory}, we explicitly store hidden states of recurrent units at different time points to extend the time span.

A vanilla choice is to take each time step into account via storing all hidden states over the whole sequence as $M = \{m_1,m_2, ..., m_{N-1}, m_N\}$, where $m_i$ denotes the $i$th hidden state in the sequence, and $N$ is the size of the memory buffer. Since contents of two consecutive images are much overlapped, it's redundant to remember each hidden state. Instead, only key states are selected. As the difference between two frames coincides with the poses, we utilize the motion distance as a metric to decide if current hidden state should be stored. 

Specifically, the current hidden state would not be put into the \textit{Memory}, unless the parallax between the current and the latest view in the slot is large enough. Here, the rotational and translational distances are utilized: 
\begin{align}
||Rot_{m_i} - Rot_{m_{i-1}}||_2 &\geq \theta_{Rot} \ ,\\
||Trans_{m_i} - Trans_{m_{i-1}}||_2 &\geq \theta_{Trans}\ . 
\end{align}
This strategy guarantees both the co-visibility of different views and the existence of global information. As both previous and new observations are gathered, the \textit{Memory} can be used to optimize previous poses.


\begin{figure}[t]
	\centering
	\subfigure[]{
		\label{fig:reweight_context}
		\begin{minipage}{0.26\textwidth}
			\centering
			\includegraphics[height=4.cm]{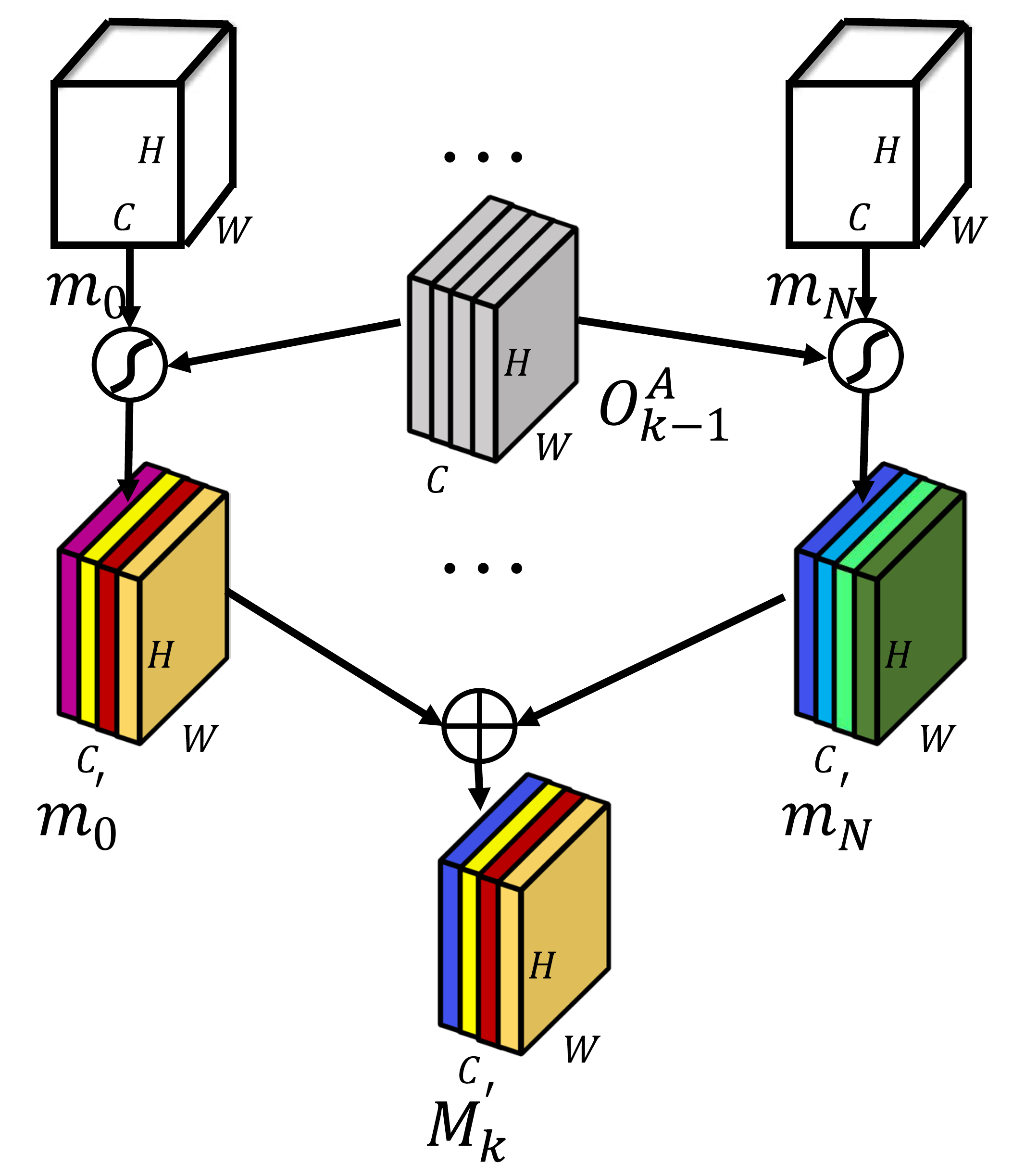}
	\end{minipage}}
	\subfigure[]{
		\label{fig:reweight_channel}
		\begin{minipage}{0.2\textwidth}
			\centering
			\includegraphics[height=4.cm]{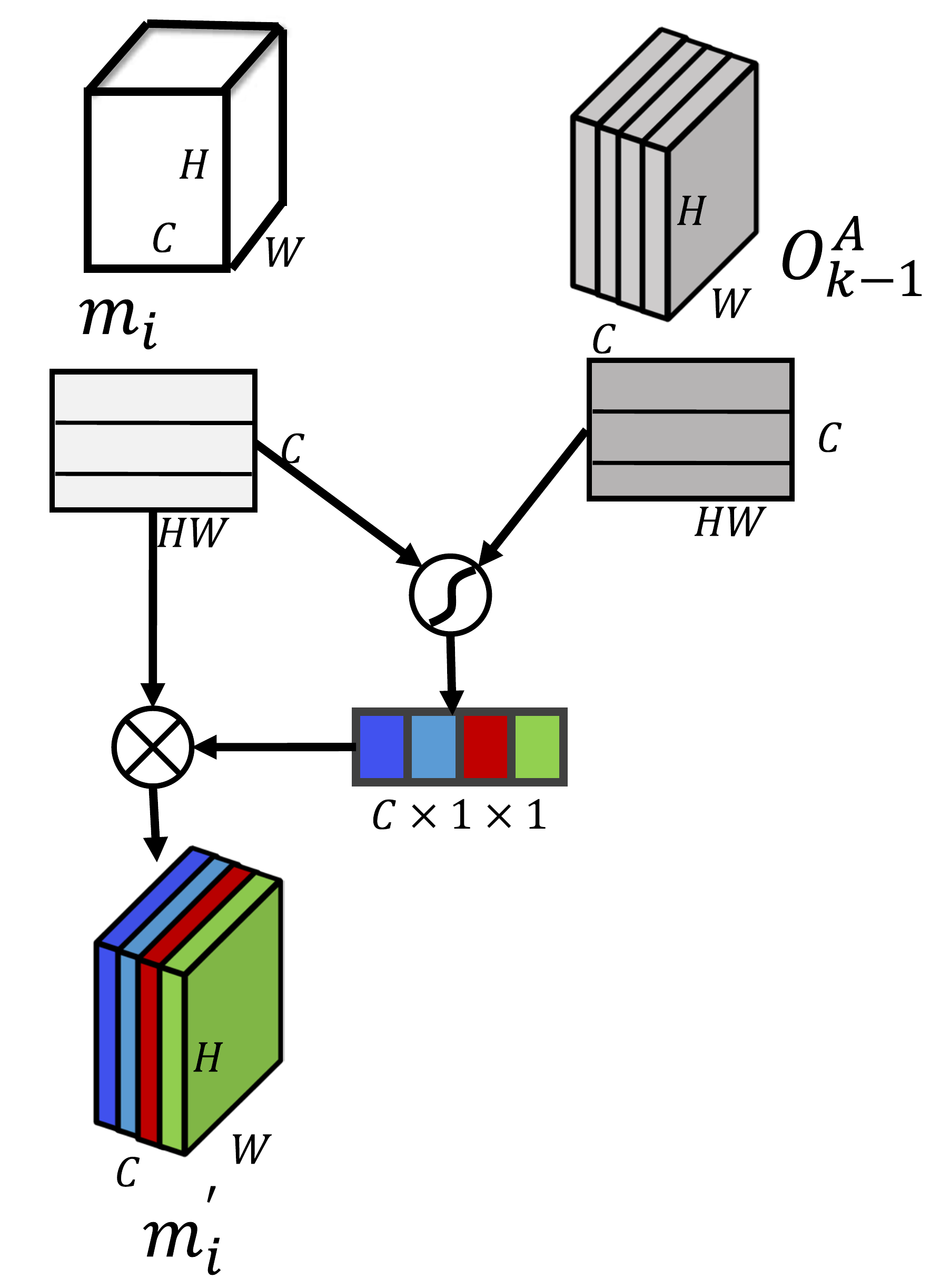}
	\end{minipage}}
	\caption{Extracting features from \textit{Memory} using the last output as guidance. We consider the correlation of both each context stored in the \textit{Memory} in (a) and every channel of the context in (b).}
	\label{fig:reweight_features}
\end{figure}

\subsection{Refining}
\label{refinement}
Once the \textit{Memory} is constructed, the \textit{Refining} module estimates the absolute pose of each view by aligning corresponding observation with the \textit{Memory}, as shown in Fig.~\ref{fig:refining}. We adopt another recurrent branch using ConvLSTM, enabling previously refined outputs passing through recurrent units to improve the next estimation, as:
\begin{align}
O_t^A, H_t^A &= \mathcal{U}^A(X_t^A, H_{t-1}^A) \ .
\end{align}
$X_t^A, O_t^A$ and $H_t^A$ are the input, output and hidden state at time $t$. $H_{t-1}^A$ denotes the hidden state at time $t-1$. The $\mathcal{U}^A$ indicates the recurrent branch for the \textit{Absolute} pose estimation. All these variables are 3D tensors to be discussed in the following sections.

\textbf{Spatial-temporal attention} Although all observations are fused and distributed in $N$ hidden states, each hidden state stored in the \textit{Memory} contributes discriminatively to different views. In order to distinguish related information, an attention mechanism is adopted. We utilize the last output $O_{t-1}^A$ as guidance, since motions between two consecutive views in a sequence are very small. 

In specific, we generate selected memories $M_{t}^{'}$ for current view $t$ with the function $\mathcal{G}$ as:
\begin{align}
	M_{t}^{'} = \mathcal{G}(O_{t-1}^A, M)\ .
\end{align}

The temporal attention aims to re-weight elements in the \textit{Memory} considering the contribution of each $m_i$ to the pose estimation of specific views. Therefore, as shown in Fig.~\ref{fig:reweight_context},  $M_{t}^{'}$ can be defined as the linear averaging of all elements in $M$ as $M_{t}^{'} = \sum_{i=1}^{N}\alpha_{i}m_i$. The $\alpha_{i} = \frac{\exp(w_{i})}{\sum_{k=1}^N\exp(w_{i})}$ denotes the normalized weight. The $w_{i} = S(O_{t-1}^A,m_i)$ is the weight computed according to the cosine similarity function denoted as $S$.

As all elements in the \textit{Memory} are formulated as 3D tensors, spatial connections are retained. In this framework, we focus on not only which element in the \textit{Memory} plays a more important role but also where each element influences the final results more significantly. We try to find corresponding co-visible contents at the feature level. Hence, we extend the attention mechanism from the temporal domain to the spatial-temporal domain incorporating an additional channel favored feature attention mechanism. Feature-map of each channel is taken as a unit and re-weighted for each view according to the last output. As shown in Fig.~\ref{fig:reweight_channel}, the process is described as:
\begin{align}
M_{t}^{'} & = \sum_{i=1}^{N}\alpha_{i}\mathcal{C}(\beta_{i1}m_{i1}, \beta_{i2}m_{i2}, ..., \beta_{iC}m_{iC}) \ .
\end{align}
The $m_{ij} \in \mathbb{R}^{H \times W}$ denotes the $j$th channel of the $i$th element in the \textit{Memory}. The $\beta_{ij}$ is the normalized weight defined on the correlation between the $j$th channel of $O_{t-1}$ and $m_i$. $\mathcal{C}$ concatenates all reweighted feature maps along the channel dimension. We calculate the cosine similarity between two vectorized feature-maps to assign the correlation weights. 

\textbf{Absolute pose estimation} The guidance is also executed on the observations encoded as high-level features to distill related visual cues, denoted as $X_t^{'}$. Both reorganized memories and observations are stacked along channels and passed through two convolutional layers with kernel size of 3 for fusion. The fused feature denoted as $X_t^{A}$ is the final input to be fed into convolutional recurrent units. Then the $\mathbb{SE}$ (3) layer calculates the absolute pose from the output $O_t^A$. Note that, through recurrent units, the hidden state propagating refined results to next time point further improves the following prediction. 

\subsection{Loss Function}
\label{loss_function}
Our model learns relative and absolute poses in the \textit{Tracking} and \textit{Refining} modules separately. Therefore, consisting of both relative and absolute pose errors, the loss functions are defined as:
\begin{align}
\mathcal{L}_{local} & = \frac{1}{t}\sum_{i=1}^{t} ||\hat{\bm{p}}_{i-1, i} - \bm{p}_{i-1, i}||_2 + k||\hat{\bm{\phi}}_{i-1, i} - \bm{\phi}_{i-1, i}||_2, \label{equ:local} \\
\mathcal{L}_{global} & = \sum_{i=1}^{t} \frac{1}{i}(||\hat{\bm{p}}_{0, i} - \bm{p}_{0, i}||_2 + k||\hat{\bm{\phi}}_{0, i} - \bm{\phi}_{0, i}||_2), \label{equ:global}\\
\mathcal{L}_{total} & = \mathcal{L}_{local} + \mathcal{L}_{global},
\end{align} 
where $\hat{\bm{p}}_{i-1, i}, \bm{p}_{i-1, i}, \hat{\bm{\phi}}_{i-1, i}$, and $\bm{\phi}_{i-1, i}$ respectively represent the predicted and ground-truth relative translations and rotations in three directions; $\hat{\bm{p}}_{0, i}, \bm{p}_{0, i}, \hat{\bm{\phi}}_{0, i} $, and $\bm{\phi}_{0, i}$ represent the predicted and ground-truth absolute translations and rotations. $\mathcal{L}_{local}, \mathcal{L}_{global}$ and $\mathcal{L}_{total}$ denote the local, global, and total losses respectively. $t$ is the current frame index in a sequence. $k$ is a fixed parameter for balancing the rotational and translational errors. 

\section{Experiments}
\label{experiment}
We first discuss the implementation details of our framework in Sec.~\ref{implementation}. Next, we compare our method with state-of-the-art approaches on the KITTI \cite{geiger2012kitti} and TUM-RGBD \cite{tum12iros} datasets in Sec.~\ref{result_kitti} and Sec.~\ref{result_tum}, respectively. Finally, an ablation study is performed in Sec.~\ref{ablation_study}.

\subsection{Implementation}
\label{implementation}
\textbf{Training} Our network takes monocular RGB image sequences as input. The image size can be arbitrary because our model has no requirement of compressing features into vectors as DeepVO \cite{wang2017deepvo} and ESP-VO \cite{ wang2018espvo}. We use 11 consecutive images to construct a sequence, yet our model can accept dynamic lengths of inputs. The parameter $k$ is set to 100 and 1 for the KITTI and TUM-RGBD dataset. The $\theta_{Rot}$ and $\theta_{Trans}$ are set to 0.005 (rad) and 0.6 (m) for the KITTI dataset. While for the TUM-RGBD dataset, the values are 0.01 (rad) and 0.01 (m). The buffer size $N$ is initialized with the sequence length, yet the buffer can be used without being fully occupied.

\textbf{Network} The encoder is pretrained on the FlyingChairs dataset \cite{dosovitskiy2015flownet}, while other parts of the network are initialized with MSRA \cite{he2015msra}. Our model is implemented by PyTorch \cite{pytorch} on an NVIDIA 1080Ti GPU. Adam \cite{Kingma2014Adam} with $\beta_1=0.9, \beta_2=0.99$ is used as the optimizer. The network is trained with a batch size of 4, a weight decay of $4 \times 10^{-4}$ for 150,000 iterations in total. The initial learning rate is set to $10^{-4}$ and reduced by half every 60,000 iterations.



\subsection{Results on the KITTI Dataset}
\label{result_kitti}
The KITTI dataset \cite{geiger2012kitti}, one of the most influential outdoor VO/SLAM benchmark datasets, is widely used in both classic \cite{mur2017orb-slam2, geiger2011stereoscan} and learning-based works \cite{zhou2017egomotion, yin2018geonet, zhan2018feature, li2018undeepvo, mahjourian2018vid2depth, wang2017deepvo, wang2018espvo}. It consists of 22 sequences captured in urban and highway environments at a relatively low sample frequency (10 fps) at the speed up to 90km/h. Seq 00-10 provide raw data with ground-truth represented as 6-DoF motion parameters considering the complicated urban environments,  while Seq 11-21 provide only raw data. In our experiments, the left RGB images are resized to 1280 x 384 for training and testing. We adopt the same train/test split as DeepVO \cite{wang2017deepvo} and GFS-VO \cite{xue2018fea} by using Seq 00, 02, 08, 09 for training and Seq 03, 04, 05, 06, 07, 10 for evaluation. 

\setlength{\tabcolsep}{3.pt}
\begin{table*}[t]
	\small
		\setlength{\abovecaptionskip}{0pt}%
	\setlength{\belowcaptionskip}{0pt}%
	\centering
		\begin{center}
			\begin{tabular}{lclclclclclclcl}
				\hline
				\hline
				& \multicolumn{14}{c}{Sequence} \\
				Method & \multicolumn{2}{c}{03} & \multicolumn{2}{c}{04} & \multicolumn{2}{c}{05} & \multicolumn{2}{c}{06} & \multicolumn{2}{c}{07} & \multicolumn{2}{c}{10} & \multicolumn{2}{c}{Avg}\\ 
				& $t_{rel}$ & $r_{rel}$ & $t_{rel}$ & $r_{rel}$ & $t_{rel}$ & $r_{rel}$ &  $t_{rel}$ & $r_{rel}$ &  $t_{rel}$ & $r_{rel}$ &  $t_{rel}$ & $r_{rel}$ &  $t_{rel}$ & $r_{rel}$   \\
				\hline
				UnDeepVO \cite{li2018undeepvo} & 5.00 & 6.17 & 5.49 & 2.13 & 3.40 & 1.50 & 6.20 & 1.98 & 3.15 & 2.48 & 10.63 & 4.65 & 5.65 & 3.15\\
				Depth-VO-Feat \cite{zhan2018feature} & 15.58 & 10.69 & 2.92 & 2.06 & 4.94 & 2.35 & 5.80 & 2.07 & 6.48 & 3.60 & 12.45 & 3.46 & 7.98 & 4.04\\
				GeoNet \cite{yin2018geonet} & 19.21 & 9.78 & 9.09 & 7.54 & 20.12 & 7.67 & 9.28 & 4.34 & 8.27 & 5.93 & 20.73 & 9.04 & 13.12 & 7.38\\
				Vid2Depth \cite{mahjourian2018vid2depth} & 27.02 & 10.39 & 18.92 & $\mathbf{1.19}$ & 51.13 & 21.86 & 58.07 & 26.83 & 51.22 & 36.64 & 21.54 & 12.54 & 37.98 & 18.24\\
				SfmLearner \cite{zhou2017egomotion} & 10.78 & 3.92 & 4.49 & 5.24 & 18.67 & 4.10 & 25.88 & 4.80 & 21.33 & 6.65 & 14.33 & 3.30 & 15.91 & 4.67\\
				DeepVO \cite{wang2017deepvo} & 8.49 & 6.89 & 7.19 & 6.97 & 2.62 & 3.61 & 5.42 & 5.82 & 3.91 & 4.60 & 8.11 & 8.83 & 5.96 & 6.12\\
				ESP-VO \cite{wang2018espvo} & 6.72 & 6.46 & 6.33 & 6.08 & 3.35 & 4.93 & 7.24 & 7.29 & 3.52 & 5.02 & 9.77 & 10.2 & 6.15 & 6.66\\
				GFS-VO-RNN \cite{xue2018fea} & 6.36 & 3.62 & 5.95 & 2.36 & 5.85 & 2.55 & 14.58 & 4.98 & 5.88 & 2.64 & 7.44 & 3.19 & 7.68 & 3.22 \\
				GFS-VO \cite{xue2018fea}  & 5.44 & 3.32 & $\mathbf{2.91}$ & 1.30 & 3.27 & 1.62 & 8.50 & 2.74 & 3.37 & 2.25 & 6.32 & 2.33 & 4.97 & 2.26\\
				
				$\mathbf{Ours}$  & $\mathbf{3.32}$ & $\mathbf{2.10}$ & 2.96 & 1.76 & $\mathbf{2.59}$ & $\mathbf{1.25}$ & $\mathbf{4.93}$ & $\mathbf{1.90}$ & $\mathbf{3.07}$ & $\mathbf{1.76}$ & $\mathbf{3.94}$ & $\mathbf{1.72}$ & $\mathbf{3.47}$ & $\mathbf{1.75}$\\
				\hline
				\hline
			\end{tabular}
		\end{center}
	
		\begin{tablenotes}
			\footnotesize
			\item $t_{rel}: $ average translational RMSE drift (\%) on length from 100, 200 to 800 m.
			\item $r_{rel}: $ average rotational RMSE drift (${}^{\circ}$/100m) on length from 100, 200 to 800 m.
		\end{tablenotes}
	\caption{Results on the KITTI dataset. DeepVO \cite{wang2017deepvo}, ESP-VO \cite{wang2018espvo}, GFS-VO \cite{xue2018fea} and our model are supervised methods trained on Seq 00, 02, 08 and 09. SfmLearner \cite{zhou2017egomotion}, GeoNet \cite{yin2018geonet}, Vid2Depth \cite{mahjourian2018vid2depth}, Depth-VO-Feat \cite{zhan2018feature}, and UndeepVO \cite{li2018undeepvo} are trained on Seq 00-08 in an unsupervised manner. The results of SfmLearner and UnDeepVO are from \cite{yang2018dvso}, while for GeoNet, Vid2Depth and Depth-VO-Feat, the poses are recovered from officially released pretrained models. The best results are highlighted.}
	\label{tab:table_kitti_00_10}
\end{table*}
\begin{figure}[t]
	\setlength{\abovecaptionskip}{0pt}%
	\setlength{\belowcaptionskip}{0pt}%
	\begin{center}
	\subfigure[Translation against path length.]{
		\begin{minipage}{0.23\textwidth}
			\centering
			\includegraphics[height=3.cm]{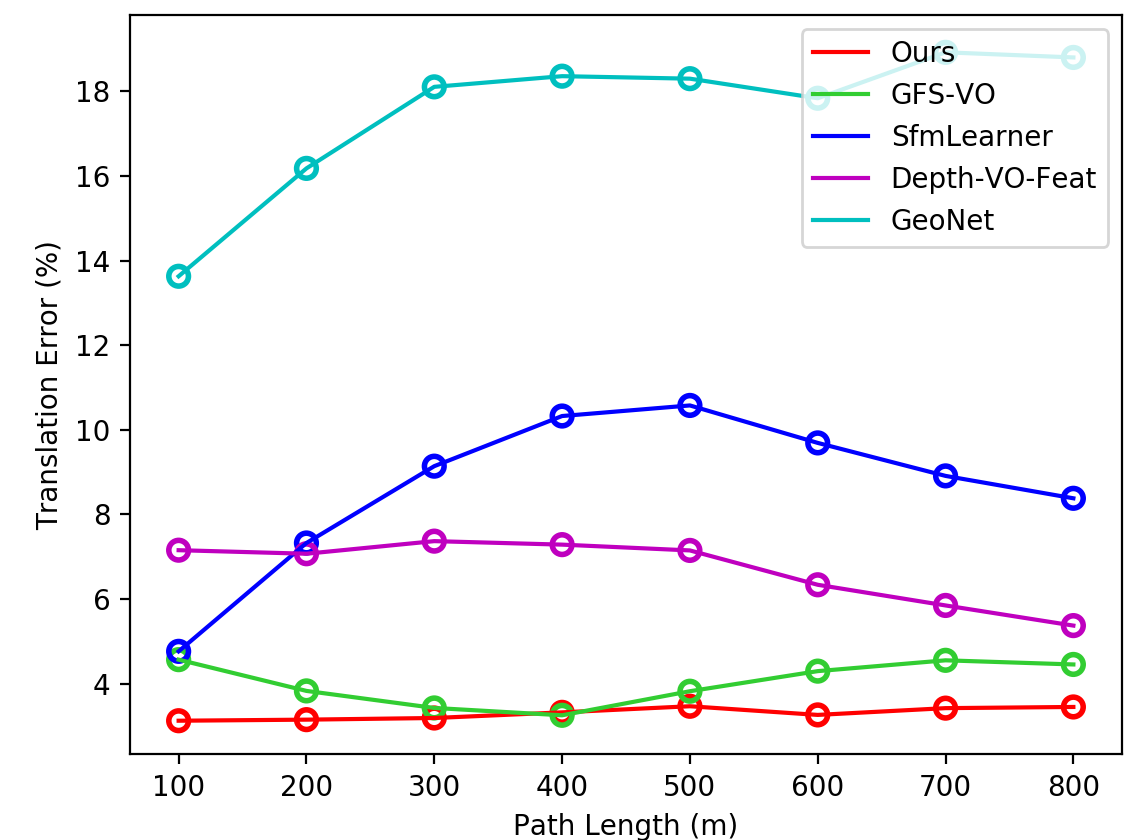}
	\end{minipage}}
	\subfigure[Rotation against path length.]{
	\begin{minipage}{0.23\textwidth}
		\centering
		\includegraphics[height=3.cm]{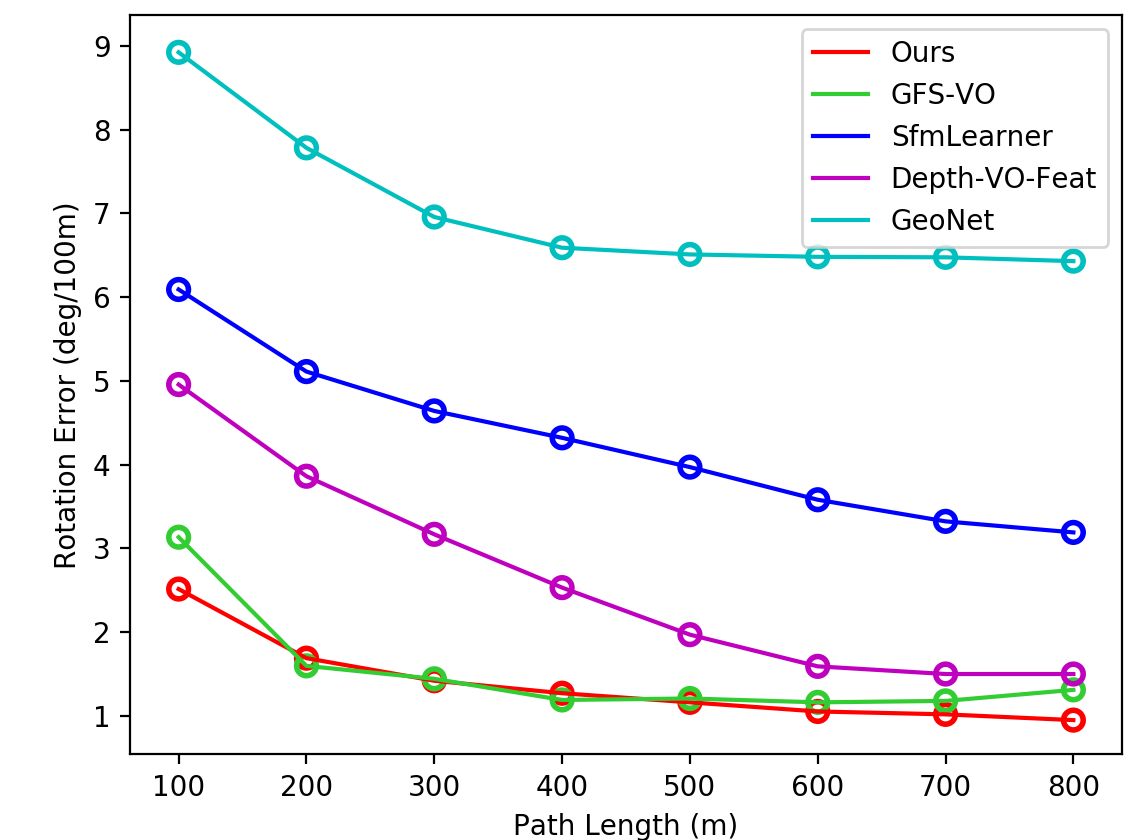}
	\end{minipage}}

	\subfigure[Translation against speed.]{
		\begin{minipage}{0.23\textwidth}
			\centering
			\includegraphics[height=3.cm]{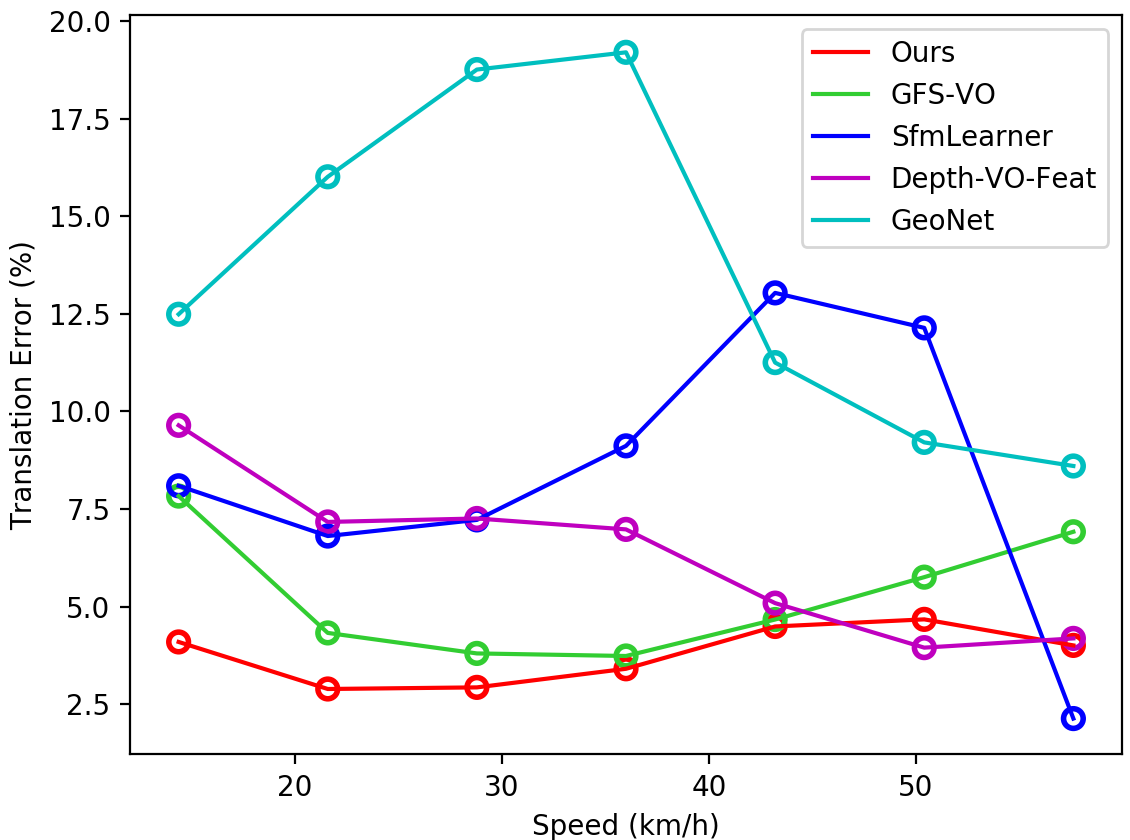}
	\end{minipage}}
	\subfigure[Rotation against speed.]{
	\begin{minipage}{0.23\textwidth}
		\centering
		\includegraphics[height=3.cm]{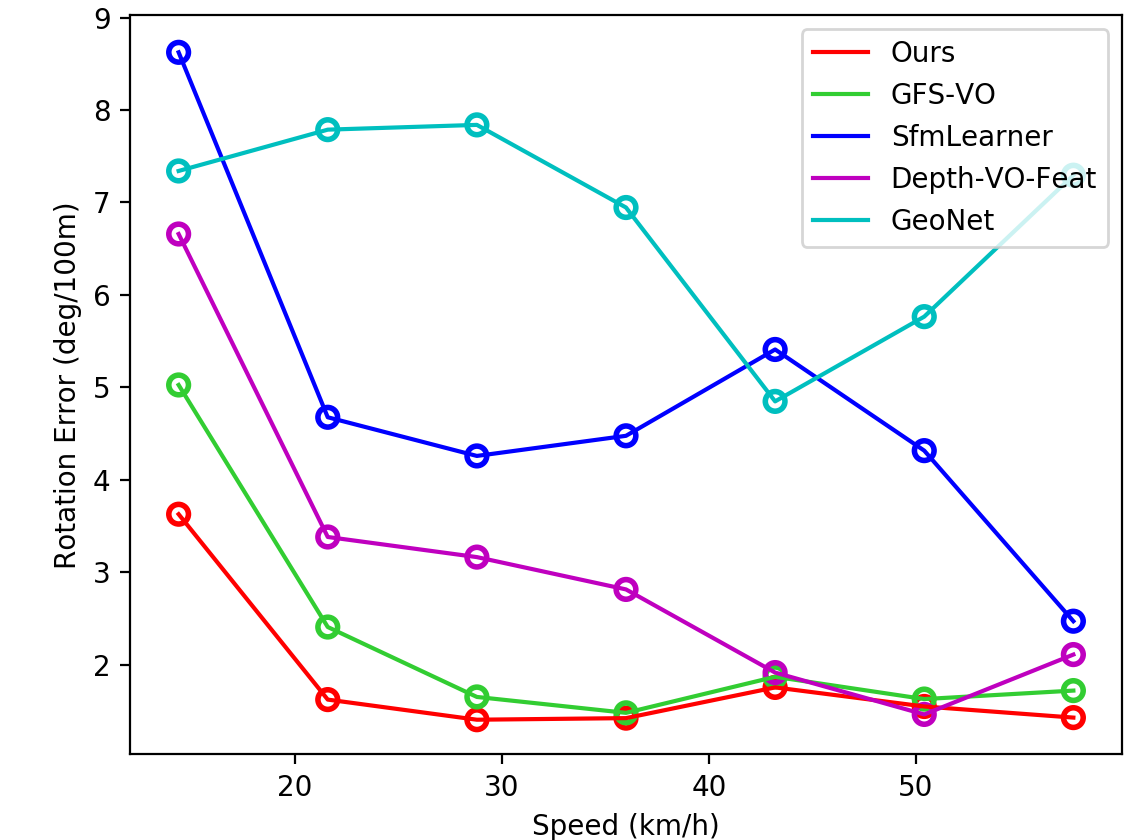}
	\end{minipage}}

	\caption{Average errors on translation and rotation against different path lengths and speeds.}
	\label{fig:trajectory_kitti_trls}
	\end{center}
	
\end{figure}
\setlength{\tabcolsep}{1.4pt}

\textbf{Baseline methods} The learning-based baselines include supervised approaches such as DeepVO \cite{wang2017deepvo}, ESP-VO \cite{wang2018espvo}, GFS-VO \cite{xue2018fea}, and unsupervised approaches such as SfmLearner \cite{zhou2017egomotion}, Depth-VO-Feat \cite{zhan2018feature}, GeoNet \cite{yin2018geonet}, Vid2Depth \cite{mahjourian2018vid2depth} and UndeepVO \cite{li2018undeepvo}. Monocular VISO2 \cite{geiger2011stereoscan} (VISO2-M) and ORB-SLAM2 \cite{mur2017orb-slam2} are used as classic baselines. The error metrics, i.e., averaged Root Mean Square Errors (RMSE) of the translational and rotational errors, are adopted for all the test sequences of the lengths ranging from 100, 200 to 800 meters.

\textbf{Comparison with learning-based methods} As shown in Table~\ref{tab:table_kitti_00_10}, our method outperforms DeepVO \cite{wang2017deepvo}, ESP-VO \cite{wang2018espvo} and GFS-VO-RNN \cite{xue2018fea} (without motion decoupling) on all of the test sequences by a large margin. Since DeepVO, ESP-VO and GFS-VO only consider historical knowledge stored in a single hidden state, error accumulates severely. The problem is partially mitigated by considering the discriminative ability of features to different motion patterns in GFS-VO, while our method is more effective.

Meanwhile, we provide the results of unsupervised approaches in Table~\ref{tab:table_kitti_00_10}. As monocular VO methods including SfmLearner \cite{zhou2017egomotion}, GeoNet \cite{yin2018geonet}, Vid2Depth \cite{mahjourian2018vid2depth} suffer from scale ambiguity, frame-to-frame motions of short sequence snippets are aligned individually with ground-truths to fix scales.  Although they achieve promising performance on sequences consisting of 5 (SfmLearner, GeoNet) or 3 (Vid2Depth) frames, they suffer from heavy error accumulation when integrating poses over the entire sequence. Benefited from stereo images in scale recovery, UnDeepVO \cite{li2018undeepvo} and Depth-VO-Feat \cite{zhan2018feature} obtain competitive results against DeepVO, ESP-VO, and GFS-VO, while our results are still much better. Note that only monocular images are used in our model. 

We further evaluate the average rotation and translation errors for different path lengths and speeds in Fig.~\ref{fig:trajectory_kitti_trls}. The accumulated errors over long path lengths are effectively mitigated by our method owing to the new information for refining previous results. Moreover, this advantage of our algorithm can also be seen in handling high speed situations. GFS-VO \cite{xue2018fea} also achieves promising rotation estimation by decoupling the motions. Unfortunately, it does not provide robust translation results.

\begin{table}[t]
	\small
	\centering
		\begin{center}
			\begin{tabular}{ccccccccc}
				\hline
				\hline
				& \multicolumn{8}{c}{Method} \\
				Seq & \multicolumn{2}{c}{Ours } & \multicolumn{2}{c}{\makecell{VISO2-M \\ \cite{geiger2011stereoscan}}} & \multicolumn{2}{c}{\makecell{ORB-SLAM2 \\ \cite{mur2017orb-slam2}}} & \multicolumn{2}{l}{ \makecell{ORB-SLAM2 \\ (LC) \cite{mur2017orb-slam2}}} \\ 
				& $t_{rel}$ & $r_{rel}$ & $t_{rel}$ & $r_{rel}$ & $t_{rel}$ & $r_{rel}$  & $t_{rel}$ & $r_{rel}$ \\
				\hline
				03 & 3.32 & 2.10 & 8.47 & 8.82 & 2.28 & 0.40 &  2.17 & 0.39  \\
				04 & 2.96 & 1.76 & 4.69 & 4.49 & 1.41 & 0.14 & 1.07 & 0.17 \\
				05 & 2.59 & 1.25 & 19.22 & 17.58 & 13.21 & 0.22 & 1.86 & 0.24  \\
				06 & 4.93 & 1.90 & 7.30 & 6.14 & 18.68 & 0.26 & 4.96 & 0.18  \\
				07 & 3.07 & 1.76 & 23.61 & 19.11 & 10.96 & 0.37 & 1.87 & 0.39 \\
				10 & 3.94 & 1.72 & 41.56 & 32.99 & 3.71 & 0.30 & 3.76 & 0.29 \\
				Avg & 3.47 & 1.75 & 17.48 & 16.52 & 8.38 & 0.28 & 2.62 & 0.28 \\
				
				\hline
				\hline
			\end{tabular}
		\end{center}
	
	\caption{Results of VISO2-M \cite{geiger2011stereoscan}, ORB-SLAM2 (with and without loop closure) \cite{mur2017orb-slam2} and our method on the KITTI dataset.}
	\label{tab:table_kitti_00_10_classic}
\end{table}

\textbf{Comparison with classic methods} The results of VISO2-M \cite{geiger2011stereoscan}, ORB-SLAM2 \cite{mur2017orb-slam2} (with and without loop closure), and our method are shown in Table~\ref{tab:table_kitti_00_10_classic}. VISO2-M is a pure monocular VO algorithm recovering frame-wise poses. ORB-SLAM2, however, is a strong baseline, because both versions utilize local bundle adjustment for jointly optimizing poses and a global map. Our model outperforms VISO2-M consistently. ORB-SLAM2 \cite{mur2017orb-slam2} achieves superior performance in terms of rotation estimation owing to the global explicit geometric constraints. However, it suffers more from error accumulation in translation on long sequences (Seq 05, 06, 07) than our approach, which is reduced by global bundle adjustment. While for short sequences (Seq 03, 04, 10), performances of the two versions and our method are very close. The small differences between the results of ORB-SLAM2 with loop close and our method suggest that global information is retained and effectively used by our novel framework.
 
\begin{figure}[t]
	\begin{center}
	\includegraphics[width=1.\linewidth]{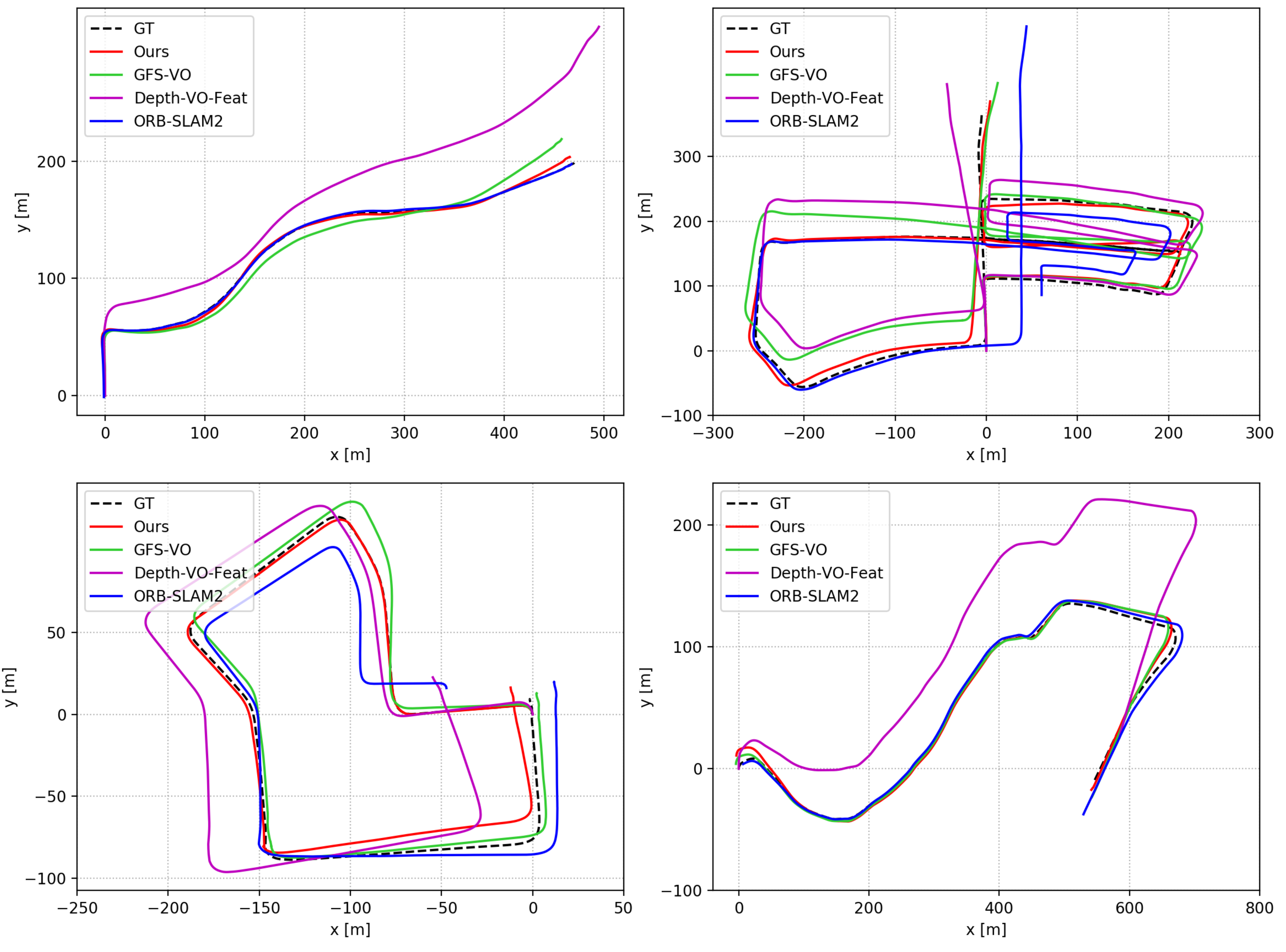}
	\end{center}
	\caption{The trajectories of ground-truth, ORB-SLAM2 \cite{mur2017orb-slam2},  Depth-VO-Feat \cite{zhan2018feature}, GFS-VO \cite{xue2018fea} and our model on Seq 03, 05, 07 and 10 (from left to right) of the KITTI benchmark.}
	\label{fig:trajectory_kitti_00_10}
\end{figure}

A visualization of the trajectories estimated by Depth-VO-Feat, GFS-VO, ORB-SLAM2 and our method is illustrated in Fig.~\ref{fig:trajectory_kitti_00_10}. Depth-VO-Feat suffers from sever error accumulation though trained on stereo images. GFS-VO and ORB-SLAM2 produces close results with our model in simple environments (Seq 03, 10), while our method outperforms them in complicated scenes (Seq 05, 07).

\subsection{Results on the TUM-RGBD Dataset}
\label{result_tum}
We test the generalization ability of our model on the TUM-RGBD dataset \cite{tum12iros}, a prevalent public benchmark used by a number of VO/SLAM algorithms \cite{mur2017orb-slam2, engel2014lsd-slam,zhou2018deeptam}. The dataset was collected by handheld cameras in indoor environments with various conditions including dynamic objects, textureless regions and abrupt motions. The dataset provides both color and depth images, while only the monocular RGB images are used in our experiments. Different from datasets captured by moving cars, motions in this benchmark contain complicated patterns due to the handheld capture mode. We select some sequences for training and others for testing (The details can be found in the supplementary material), and evaluate the performance in both regular and challenging conditions using the averaged Absolute Trajectory Errors (ATE). 

\begin{figure*}[t]
	\centering
		\begin{minipage}{1.\textwidth}
			\centering
			\includegraphics[height=6.5cm]{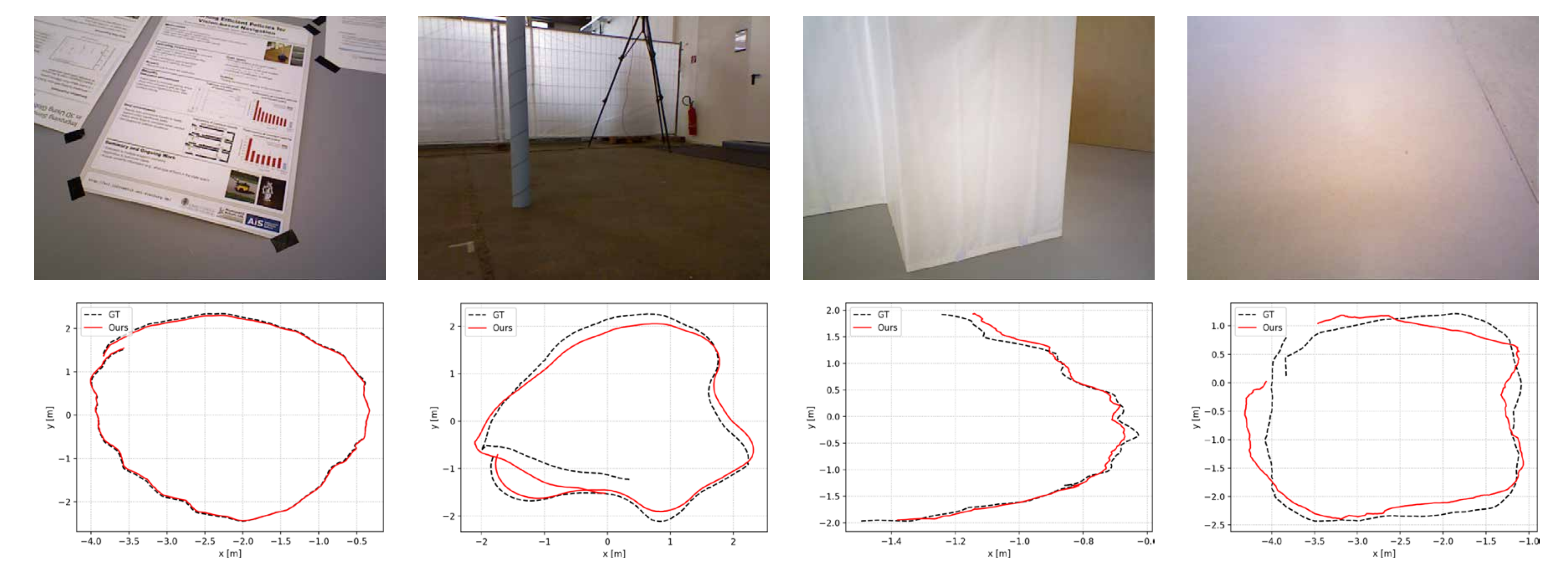}
			\label{fig:trajectory_tum1}
	\end{minipage}

	\caption{The raw images (top) and trajectories (bottom) recovered by our method on the TUM-RGBD dataset \cite{tum12iros} (from left to right: fr3/str\_tex\_far, fr2/poineer\_360, fr3/str\_ntex\_far, fr3/nstr\_ntex\_near\_loop). Trajectories are aligned with ground-truths for scale recovery. }
	\label{fig:trajectory_tum}
\end{figure*}

\setlength{\tabcolsep}{3.4pt}
\begin{table*}[t]
	\small
	\centering
		\begin{center}
			\begin{tabular}{ccc||cc|cccc}
				\hline
				\hline
				Sequence & \makecell{Desc. \\ str/tex/abrupt motion}& Frames & \makecell{ ORB-SLAM2 \\ \cite{mur2017orb-slam2}}  & \makecell{ DSO \\ \cite{engel2018dso}} & \makecell{Ours \\ (tracking)} & \makecell{Ours \\  (w/o temp atten)} & \makecell{Ours \\ (w/o spat atten)} & \makecell{Ours} \\
				\hline
				fr2/desk & Y/Y/N & 2965 & \textbf{0.041} & X  & 0.183  & 0.164 & 0.159 & 0.153 \\
				fr2/360\_kidnap & Y/Y/N & 1431 & \textbf{0.184} & 0.197  & 0.313  & 0.225 & 0.224 & 0.208 \\
				fr2/pioneer\_360  & Y/Y/Y & 1225 & X  & X  & 0.241  & 0.1338 & 0.076 & \textbf{0.056} \\
				fr2/pioneer\_slam3 & Y/Y/Y & 2544 & X & 0.737  & 0.149  & 0.1065 & 0.085 & \textbf{0.070}\\
				fr2/large\_cabinet & Y/N/N & 1011 & X & X  & 0.193  & 0.193 & 0.177 & \textbf{0.172} \\
				fr3/sitting\_static & Y/Y/N & 707 & X & 0.082  & 0.017  & 0.018 & 0.017 & \textbf{0.015} \\
				fr3/nstr\_ntex\_near\_loop & N/N/N & 1125 & X  & X  & 0.371 & 0.195 & 0.157 & \textbf{0.123} \\ 
				fr3/nstr\_tex\_near\_loop & N/Y/N & 1682 & 0.057 & 0.093  & 0.046  & 0.011 & 0.010 & \textbf{0.007} \\ 
				fr3/str\_ntex\_far & Y/N/N & 814 &X & 0.543  & 0.069  & 0.047 & 0.039 & \textbf{0.035} \\ 
				fr3/str\_tex\_far & Y/Y/N & 938 & \textbf{0.018} & 0.040  & 0.080  & 0.049 & 0.046 & 0.042 \\

				\hline
				\hline
			\end{tabular}	
		\end{center}
	\caption{Evaluation on the TUM-RGBD dataset \cite{tum12iros}. The values describe the translational RMSE in [m/s]. Results of ORB-SLAM2 \cite{mur2017orb-slam2} and DSO \cite{engel2018dso} are generated from the officially released source code with recommended parameters. \textbf{Ours (tracking)} is a network which contains only the tracking component. \textbf{Ours (w/o temp atten)} indicates the model averaging the all memories as input without temporal attention. \textbf{Ours (w/o spat atten)} is the model removing the spatial attention yet retaining the temporal attention.}
	\label{table:tum}
\end{table*}

\textbf{Comparison with classic methods} Since few monocular learning-based VO algorithms have attempted to handle complicated motions recorded by handheld cameras, we alternatively compare our approach against current state-of-the-art classic methods including ORB-SLAM2 \cite{mur2017orb-slam2} and DSO \cite{engel2018dso}.  As shown in Table~\ref{table:tum}, they yield promising results on scenes with rich textures (fr2/desk, fr2/360\_kidnap, fr3/sitting\_static, fr3/nstr\_tex\_near\_loop, fr3/str\_tex\_far), yet our results are comparable.

As ORB-SLAM2\cite{mur2017orb-slam2} relies on ORB \cite{orb2011} features to establish correspondences, it fails in scenes without rich textures (fr3/nstr\_ntex\_near\_loop, fr3/str\_ntex\_far, fr2/large\_cabinet). Utilizing pixels with large gradients for tracking, DSO \cite{engel2018dso} works well in scenes with structures or edges (fr3/str\_ntex\_far, fr3/str\_tex\_far). It cannot achieve good performance when textures are insufficient. Both ORB-SLAM2 and DSO can hardly work in scenes without texture and structure (fr2/large\_cabinet, fr3/nstr\_ntex\_near\_loop) and tend to fail when facing abrupt motions (fr2\_pioneer\_360, fr2/\_pioneer\_slam3). In contrast, our method is capable of dealing with these challenges owing to the ability of deep learning in extracting high-level features, and the efficacy of our proposal for error reduction. A visualization of trajectories is shown in Fig.~\ref{fig:trajectory_tum}.

\subsection{Ablation Study}
\label{ablation_study}
Table~\ref{table:tum} also shows an ablation study illustrating the importance of each component in our framework. The baseline is our model removing the \textit{Memory} and \textit{Refining} modules, similar to \cite{wang2017deepvo, wang2018espvo, xue2018fea}. The \textit{Tracking} model works poorly in both regular and challenging conditions, because historical knowledge in a single hidden state is inefficient to reduce accumulated errors. Fortunately, the \textit{Memory} component mitigates the problem by explicitly introducing more global information and considerably improves results of the \textit{Tracking} model both on regular and challenging sequences. 

We further test the spatial-temporal attention strategy adopted for selecting features from memories and observations by removing the \textit{temporal attention} and \textit{spatial attention} progressively. We observe that both of the two attention techniques are crucial to improve the results, especially in challenging conditions (fr2/pioneer\_360, fr2/pioneer\_slam3, fr3/nstr\_ntex\_near\_loop). 


\section{Conclusion}
\label{conclusion}
In this paper, we present a novel framework for learning monocular visual odometry in an end-to-end fashion. In the framework, we incorporate another two helpful components called \textit{Memory} and \textit{Refining}, which focus on introducing more global information and ameliorating previous results with these information respectively. We utilize an adaptive and efficient selection strategy to construct the \textit{Memory}. Besides, a spatial-temporal attention mechanism is employed for feature selection when recovering the absolute poses in the \textit{Refining} module. The refined results propagating information through recurrent units, further improve the following estimation. Experiments demonstrate that our model outperforms previous learning-based monocular VO methods and gives competitive results against classic VO approaches on the KITTI and TUM-RGBD benchmarks respectively. Moreover, our model obtains outstanding results under challenging conditions including texture-less regions and abrupt motions, where classic methods tend to fail. 

In the future, we consider to extend the work to a full SLAM system consisting tracking, mapping and global optimization. Moreover, auxiliary information, such as IMU and GPS data will also be introduced to enhance the system.  

\section*{Acknowledgement}
The work is supported by the National Key Research and Development Program of China (2017YFB1002601) and National Natural Science Foundation of China (61632003, 61771026).

\clearpage
{
	\small
	\bibliographystyle{ieee}
	\bibliography{VOMachine_final_with_supply}
}

\begin{abstract}
	In the supplementary material, we provide additional technical details and results for our paper.  We first give the details of the training/testing splits of TUM-RGBD dataset used in our experiments. Next, we test the influence of the length of the training sequence on the performance of our model. Finally, we test the generalization ability of our network on the extra sequences of the KITTI benchmark.
\end{abstract}

\section*{Training Data in the TUM-RGBD dataset}
\label{sec:training_data_tum}
We select 19 sequences for training (see Table~\ref{table:tum_training}) and 10 sequences for testing (see Table~\ref{table:tum_testing}). Both the training and testing sequences include various conditions such as regular scenes, textureless regions, or abrupt motions. This dataset was collected by handheld cameras at up to 30 fps, resulting in much overlap between consecutive frames. Training our model directly on these raw sequences will lead to over-fitting. To cope with problem, we randomly select 16578 short sequences as the training sequences, yet the overlapped frames are reduced. 

\section*{The Influence of Sequence Length}
\label{sec:sequence_length}
In this section, We test the influence of sequence length on the results. Theoretically, the better performance can be achieved by our model when given more frames. We compare the results with sequences constructing of 5, 7, 9, and 11 frames on the KITTI (see Table~\ref{tab:table_kitti_00_10_frames}) and the  TUM-RGBD dataset (see Table~\ref{table:tum_frames_full}) respectively. 

\begin{figure*}[t]
	\centering
	\subfigure[Seq 03]{
		\begin{minipage}{0.45\textwidth}
			\centering
			\includegraphics[height=5.cm]{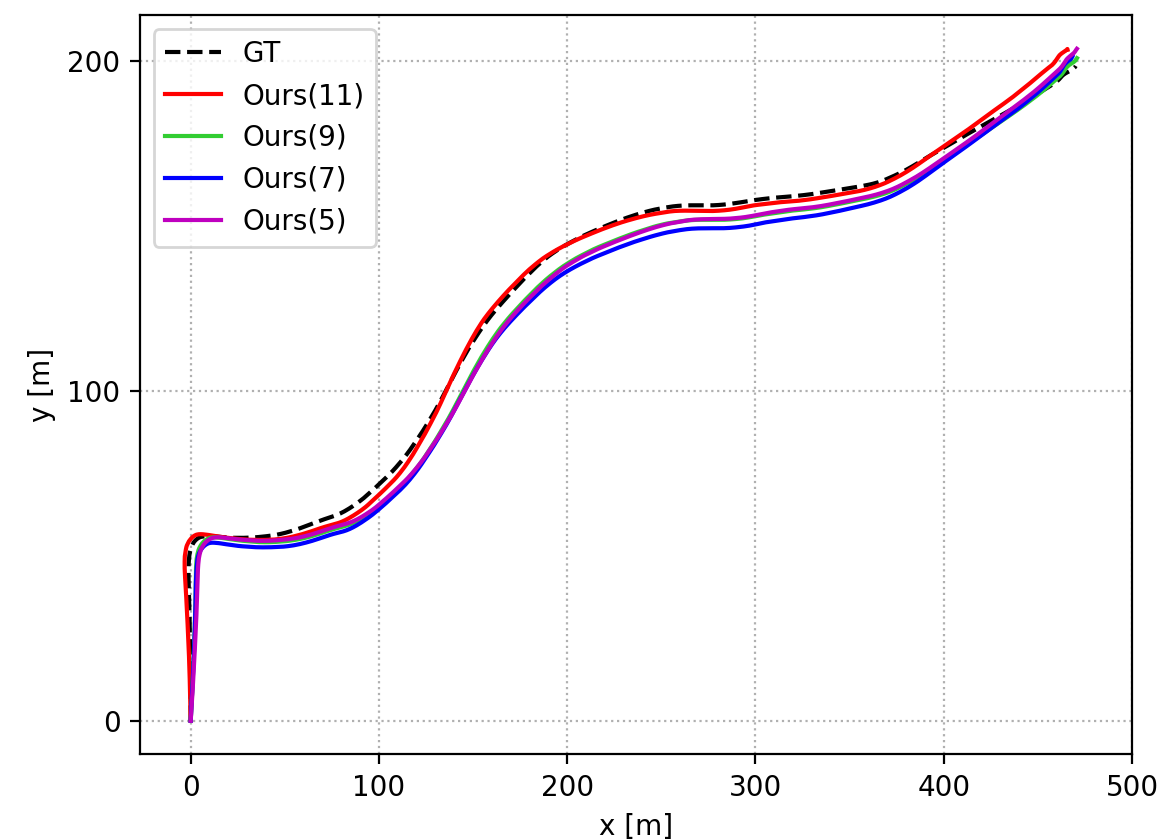}
	\end{minipage}}
	\subfigure[Seq 05]{
	\begin{minipage}{0.45\textwidth}
		\centering
		\includegraphics[height=5.cm]{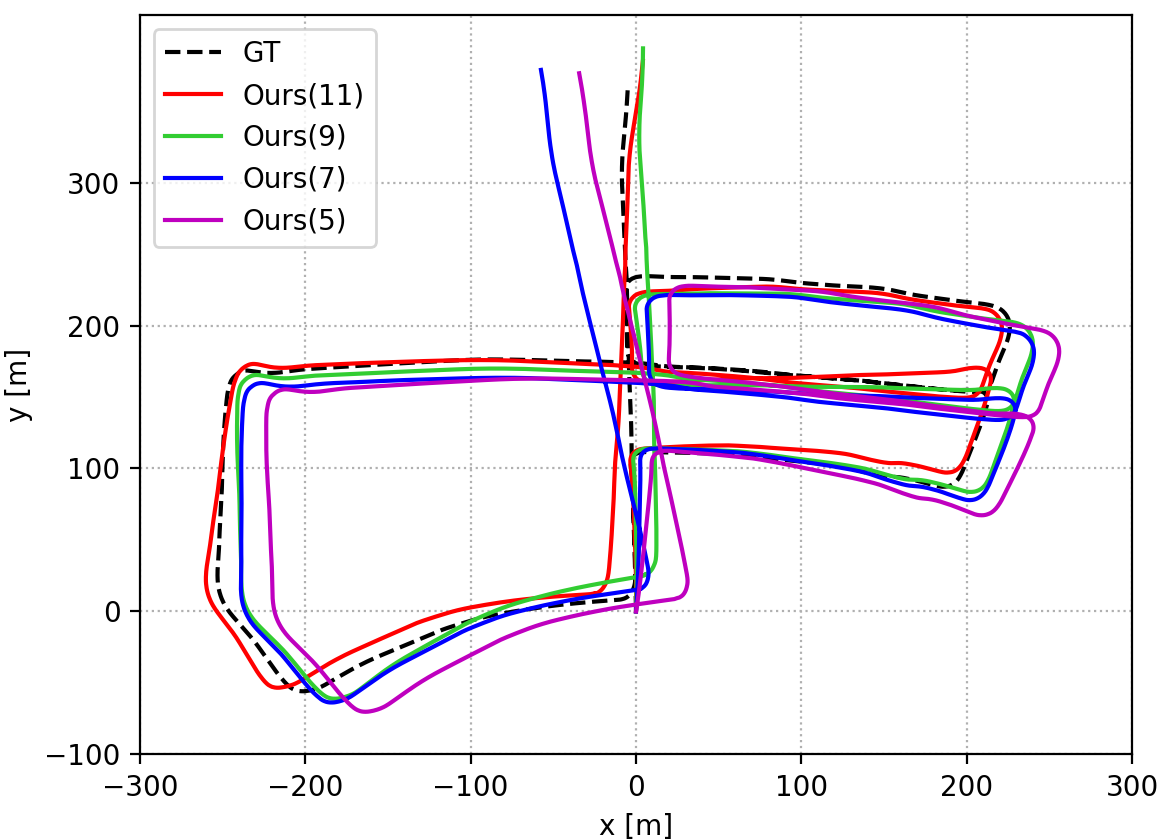}
	\end{minipage}}
	\subfigure[Seq 07]{
	\begin{minipage}{0.45\textwidth}
		\centering
		\includegraphics[height=5.cm]{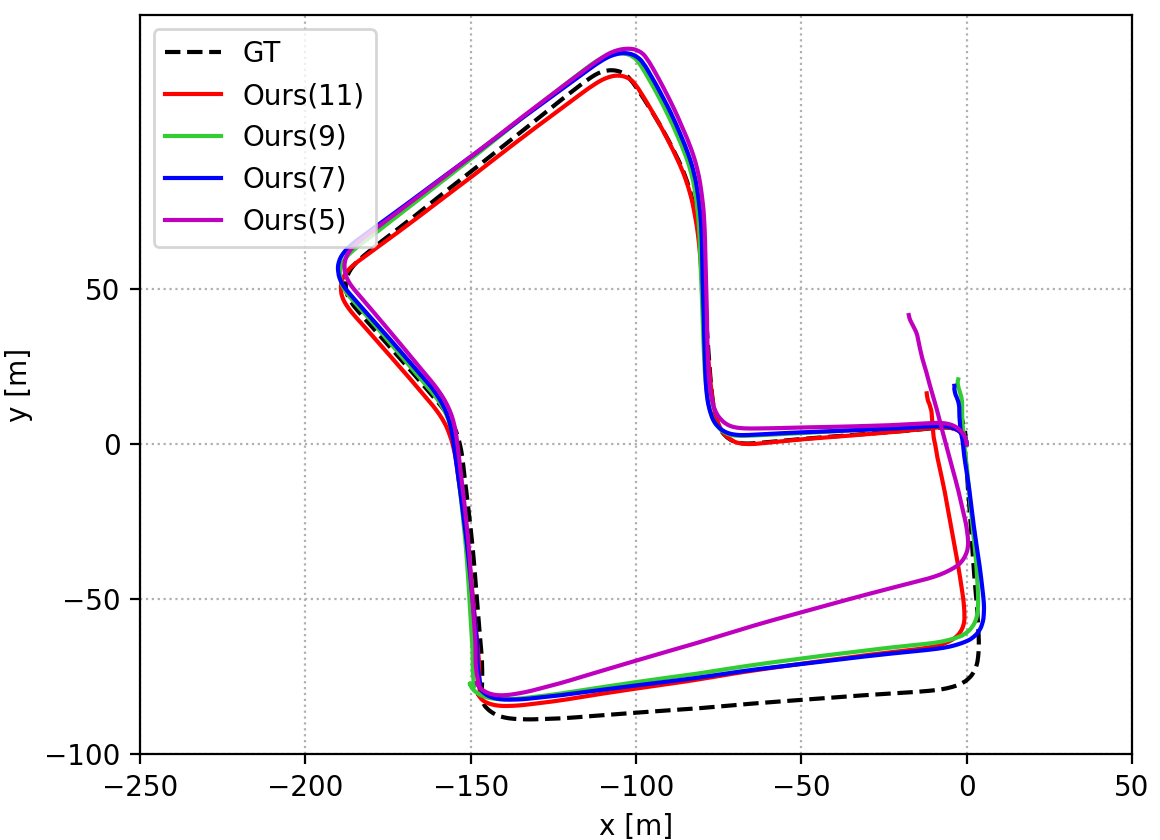}
	\end{minipage}}
	\subfigure[Seq 10]{
	\begin{minipage}{0.45\textwidth}
		\centering
		\includegraphics[height=5.cm]{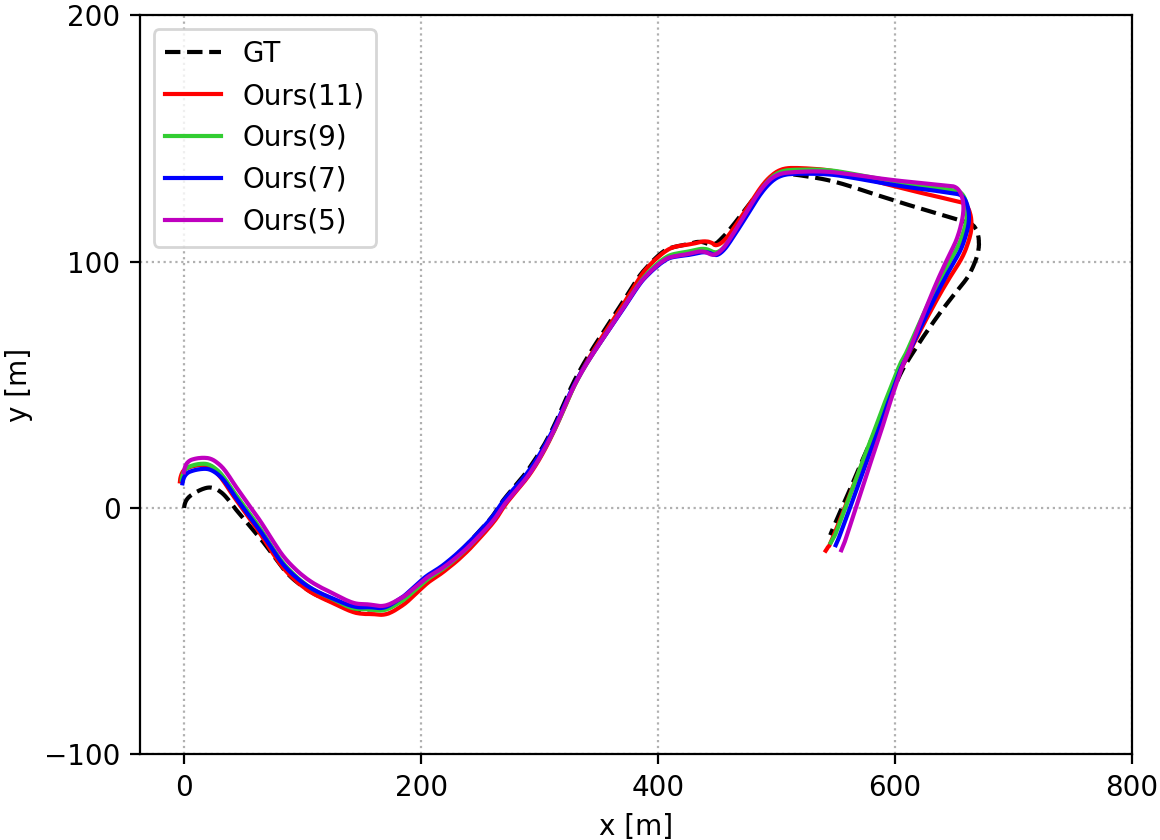}
	\end{minipage}}
	\caption{The trajectories of models trained on sequences with 5, 7, 9, and 11 frames.}
	\label{fig:trajectory_kitti_frames}
\end{figure*}

\begin{figure*}[t]
	\begin{center}
		\includegraphics[width=.85\linewidth]{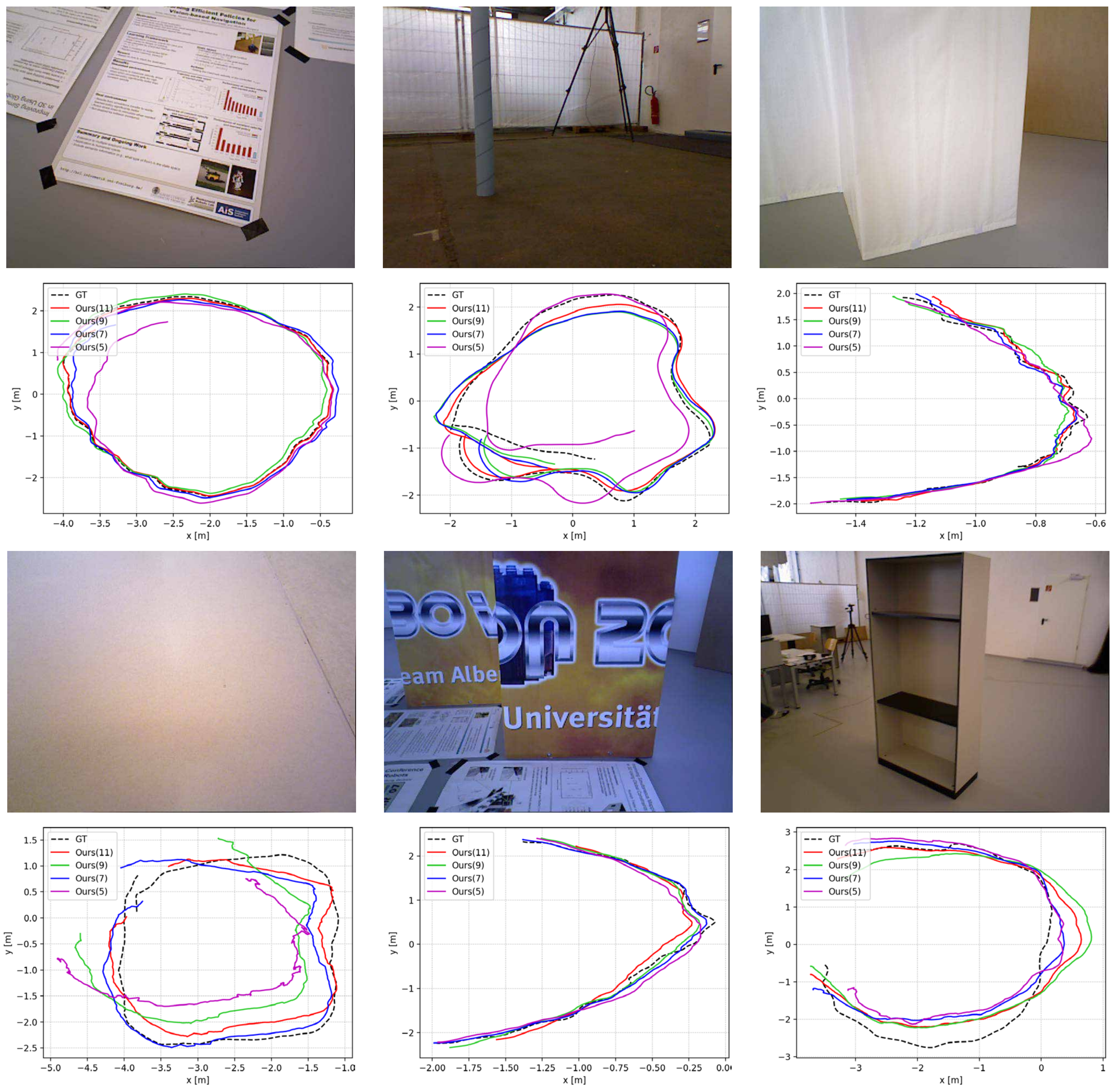}
	\end{center}
	\caption{Trajectories of our model trained with sequences consisting of different frames on the TUM-RGBD dataset \cite{tum12iros} (from left to right: fr3/nstr\_tex\_near\_loop, fr2/pioneer\_360, fr3/str\_ntex\_far, fr3/nstr\_ntex\_near\_loop, fr3/str\_tex\_far, fr3/large\_cabinet).}
	\label{fig:trajectory_tum_frames}
\end{figure*}

As we can see from Table~\ref{tab:table_kitti_00_10_frames} and \ref{table:tum_frames_full}, results of our model are improved considerably by introducing more frames. Since our model preserves information explicitly over the whole sequence and refines previous outputs with new observations, the performance can be intuitively improved by giving more observations. Fig.~\ref{fig:trajectory_kitti_frames} and \ref{fig:trajectory_tum_frames} illustrate the qualitative comparison.

\begin{table}[t]
	\small
	\centering
		\begin{center}
			\begin{tabular}{ccc}
				\hline
				\hline
				Sequence & \makecell{Desc. \\ str/tex/brute motion} & Frames \\
				\hline
				fr1/360 & Y/Y/N & 756 \\
				fr1/floor & Y/N/N & 1242 \\
				fr1/room & Y/Y/N & 1362 \\
				fr1/desk & Y/Y/Y & 613 \\
				fr1/desk2 & Y/Y/Y & 640 \\
				fr2/xyz & Y/Y/N & 3669 \\
				fr1/plant & N/Y/N & 1141 \\
				fr1/teddy & N/N/N & 1419 \\
				fr1/coke & N/Y/N & 2521 \\
				fr3/teddy & N/Y/N & 2409 \\
				fr2/flowerbouquet & Y/N/N & 2972 \\
				fr3/sitting\_xyz & Y/Y/N & 1261 \\
				fr1/sitting\_helfsphere & Y/Y/N & 1110 \\
				fr2/pioneer\_slam & Y/Y/Y & 2921 \\
				fr2/pioneer\_slam2 & Y/Y/Y & 2113 \\
				fr3/nstr\_ntex\_far & N/N/N & 474 \\
				fr3/nstr\_tex\_far & N/Y/N & 465 \\
				fr3/str\_ntex\_near & Y/N/N & 1082 \\
				fr3/str\_tex\_near & Y/Y/N & 1099 \\
				\hline
				\hline
			\end{tabular}	
		\end{center}
	\caption{Training sequences used in the TUM-RGBD dataset \cite{tum12iros}.}
	\label{table:tum_training}
\end{table}
\begin{table}[htbp]
	\small
	\centering
		\begin{center}
			\begin{tabular}{ccc}
				\hline
				\hline
				Sequence & \makecell{Desc. \\ str/tex/brute motion} & Frames \\
				\hline
				fr2/desk & Y/Y/N & 2965 \\
				fr2/360\_kidnap & Y/Y/N & 1431 \\
				fr2/pioneer\_360 & Y/Y/Y & 1225 \\
				fr2/pioneer\_slam3 & Y/Y/Y & 2113 \\
				fr2/large\_cabinet & Y/N/N & 1011 \\
				fr3/sitting\_static & Y/Y/N & 707 \\
				fr3/nstr\_ntex\_near\_loop & N/N/N & 1125 \\
				fr3/nstr\_tex\_near\_loop & N/Y/N & 1682 \\
				fr3/str\_ntex\_far & Y/N/N & 814 \\
				fr3/str\_tex\_far & Y/Y/N & 938 \\
				\hline
				\hline
			\end{tabular}	
		\end{center}
	\caption{Training sequences in the TUM-RGBD dataset \cite{tum12iros}.}
	\label{table:tum_testing}
\end{table}

\setlength{\tabcolsep}{3.pt}
\begin{table*}[t]
	
	\centering
		\begin{center}
			\begin{tabular}{lclclclclclclcl}
				\hline
				\hline
				& \multicolumn{14}{c}{Sequence} \\
				Method & \multicolumn{2}{c}{03} & \multicolumn{2}{c}{04} & \multicolumn{2}{c}{05} & \multicolumn{2}{c}{06} & \multicolumn{2}{c}{07} & \multicolumn{2}{c}{10} & \multicolumn{2}{c}{Avg}\\ 
				& $t_{rel}$ & $r_{rel}$ & $t_{rel}$ & $r_{rel}$ & $t_{rel}$ & $r_{rel}$ &  $t_{rel}$ & $r_{rel}$ &  $t_{rel}$ & $r_{rel}$ &  $t_{rel}$ & $r_{rel}$ &  $t_{rel}$ & $r_{rel}$   \\
				\hline
				\textbf{5 frames}   & 4.01 & 2.82 & 3.31 & 2.28 & 3.54 & 1.69 & 7.27 & 2.61 & 5.67 & 3.35 & 5.25 & 3.16 & 4.84 & 2.65\\
				
				\textbf{7 frames}  & 3.83 & 2.81 & 3.34 & 2.51 & 3.33 & 1.64 & 6.56 & 2.32 & $\mathbf{2.60}$ & $\mathbf{1.71}$ & 5.02 & 2.77 & 4.11 & 2.29\\
				
				\textbf{9 frames}  & 3.34 & 2.16 & 3.18 & $\mathbf{1.46}$ & 3.31 & 1.51 & 6.30 & 2.08 & 3.24 & 1.97 & 4.16 & 2.16 & 3.92 & 1.89\\
				
				\textbf{11 frames} & $\mathbf{3.32}$ & $\mathbf{2.10}$ & $\mathbf{2.96}$ & 1.76 & $\mathbf{2.59}$ & $\mathbf{1.25}$ & $\mathbf{4.93}$ & $\mathbf{1.90}$ & 3.07 & 1.76 & $\mathbf{3.94}$ & $\mathbf{1.72}$ & $\mathbf{3.47}$ & $\mathbf{1.75}$\\
				\hline
				\hline
			\end{tabular}
		\end{center}
	
		\begin{tablenotes}
			\footnotesize
			\item $t_{rel}: $ average translational RMSE drift (\%) on length from 100, 200 to 800 m.
			\item $r_{rel}: $ average rotational RMSE drift (${}^{\circ}$/100m) on length from 100, 200 to 800 m.
		\end{tablenotes}
	\caption{Results of our method using different lengths of sequences on the KITTI dataset. The best results are highlighted.}
	\label{tab:table_kitti_00_10_frames}
\end{table*}

\setlength{\tabcolsep}{3.4pt}
\begin{table*}[t]
	\centering
		\begin{center}
			\begin{tabular}{ccc||cccccc}
				\hline
				\hline
				Sequence & \makecell{Desc. \\ str/tex/abrupt motion}& Frames & \makecell{5 frames}  & \makecell{ 7 frames} & \makecell{9 frames} & \makecell{11 frames}\\
				\hline
				fr2/desk & Y/Y/N & 2965  & 0.230  & 0.177 & 0.158 & \textbf{0.153} \\
				fr2/360\_kidnap & Y/Y/N & 1431  & 0.238  & 0.228 & 0.223 & \textbf{0.208} \\
				fr2/pioneer\_360  & Y/Y/Y & 1225 & 0.106  & \textbf{0.054} & 0.062 & 0.056 \\
				fr2/pioneer\_slam3 & Y/Y/Y & 2544 & 0.105  & 0.073 & 0.072 & \textbf{0.070}\\
				fr2/large\_cabinet & Y/N/N & 1011 & 0.201  & \textbf{0.168} & 0.175 & 0.172 \\
				fr3/sitting\_static & Y/Y/N & 707 & 0.017  & 0.015 & 0.015 &\textbf{0.015} \\
				fr3/nstr\_ntex\_near\_loop & N/N/N & 1125 & 0.237 & 0.123 & 0.127 & \textbf{0.123} \\ 
				fr3/nstr\_tex\_near\_loop & N/Y/N & 1682   & 0.025  & 0.017 & 0.014 & \textbf{0.007} \\ 
				fr3/str\_ntex\_far & Y/N/N & 814 & 0.046  & 0.037 & 0.044 & \textbf{0.035} \\ 
				fr3/str\_tex\_far & Y/Y/N & 938 & 0.057  & 0.046 & 0.046 & \textbf{0.042} \\ 
				\hline
				\hline
			\end{tabular}	
		\end{center}
	\caption{Evaluation on the TUM-RGBD dataset \cite{tum12iros}. The values describe the translational RMSE in [m/s]. Results of our model with the \textit{Tracking}, \textit{Memory} and \textit{Refining} components on sequence lengths of 5, 7, 9,11. The best results are highlighted.}
	\label{table:tum_frames_full}
\end{table*}

\section*{Generalization}
\label{sec:generalization}
We tentatively test the generalization ability of our model on Seq 11-19 of the KITTI dataset \cite{geiger2012kitti}. Since the ground-truths of these sequences are unavailable, similar with GFS-VO, we utilize the results of stereo VISO2 (VISO2-S) \cite{geiger2011stereoscan} as references.
This time, our model is trained on the Seq 00-10 as GFS-VO \cite{xue2018fea} to avoid over-fitting and maximize the ability of generalization. 

\begin{figure*}[t]
	\centering
	\subfigure[Seq 11]{
		\begin{minipage}{0.32\textwidth}
			\centering
			\includegraphics[height=4.2cm]{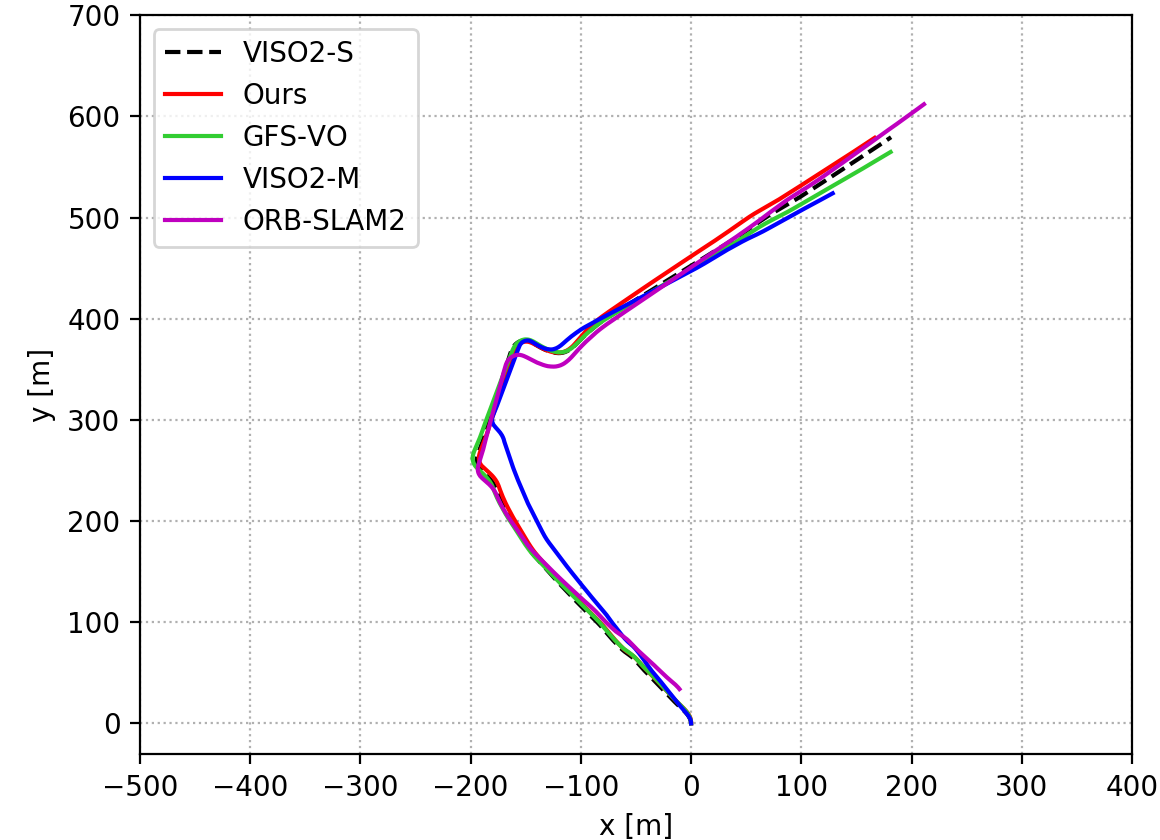}
			\label{fig:trajectory_kitti_11}
	\end{minipage}}
	\subfigure[Seq 12]{
		\begin{minipage}{0.32\textwidth}
			\centering
			\includegraphics[height=4.2cm]{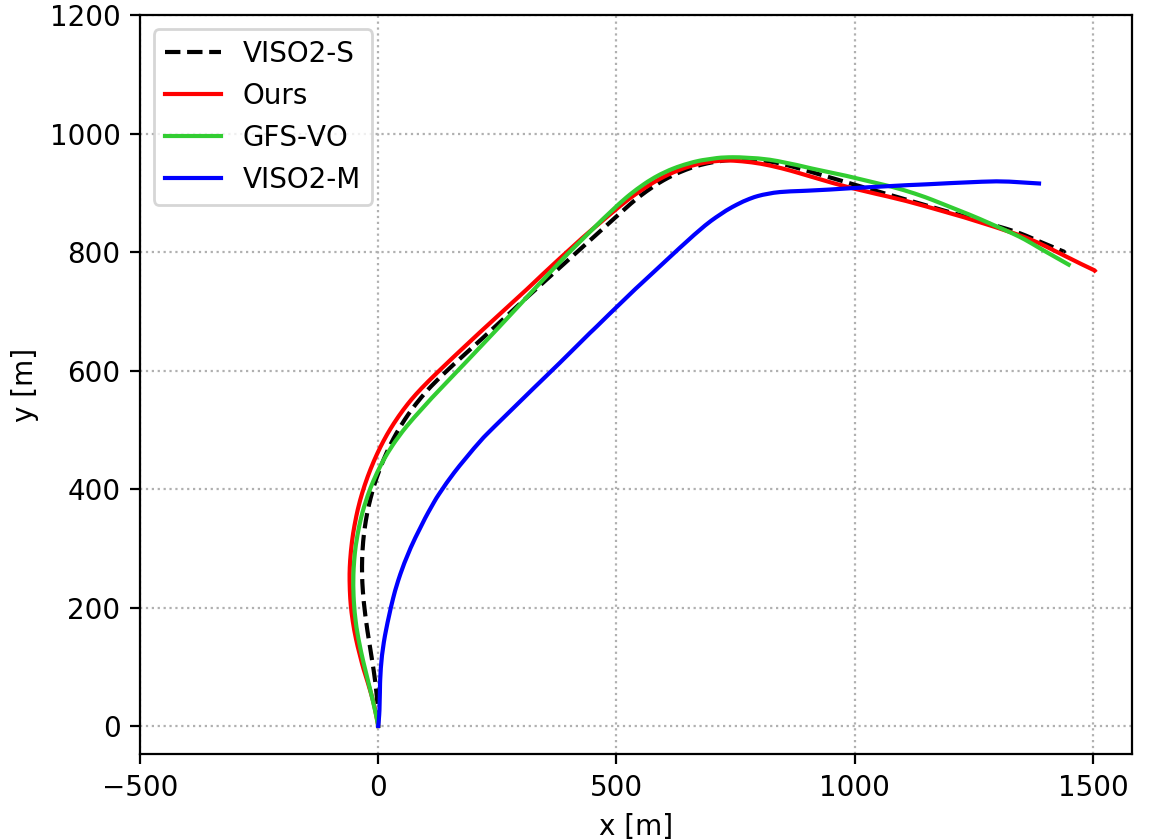}
			\label{fig:trajectory_kitti_12}
	\end{minipage}}
	\subfigure[Seq 13]{
	\begin{minipage}{0.32\textwidth}
		\centering
		\includegraphics[height=4.2cm]{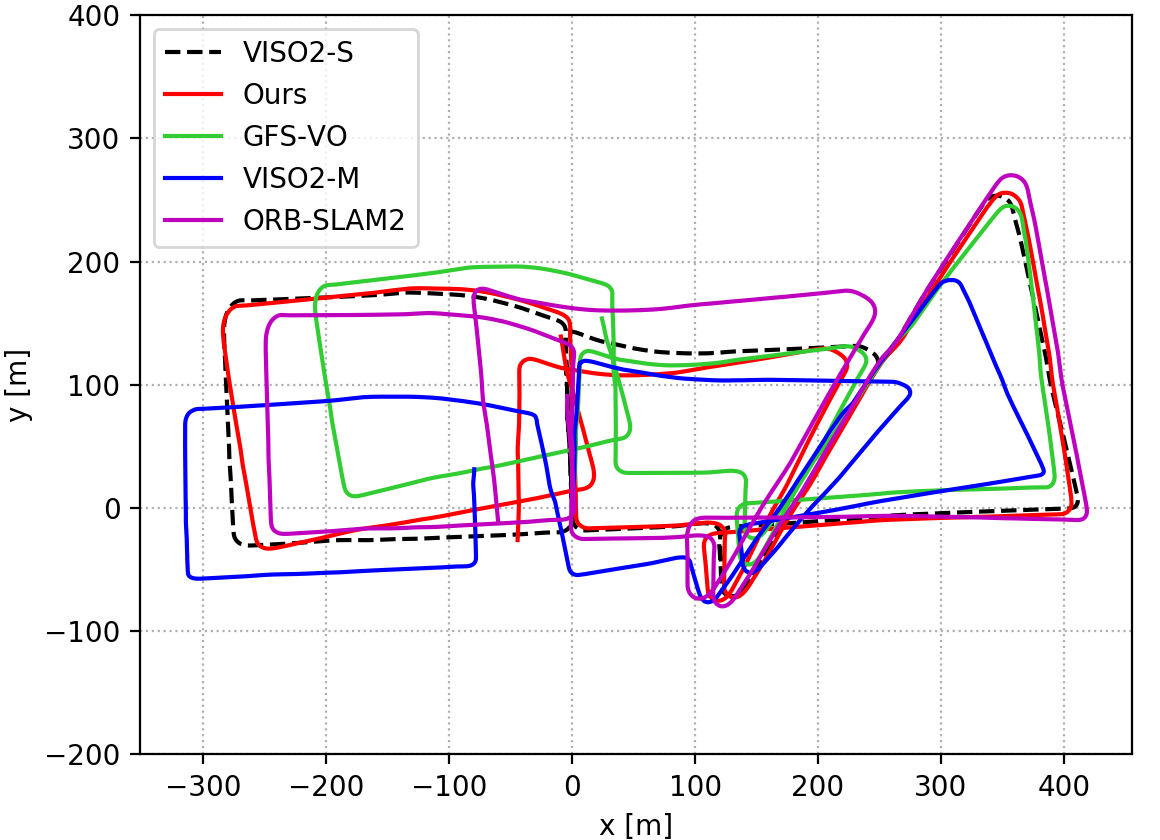}
		\label{fig:trajectory_kitti_13}
	\end{minipage}}
	\subfigure[Seq 14]{
	\begin{minipage}{0.32\textwidth}
		\centering
		\includegraphics[height=4.2cm]{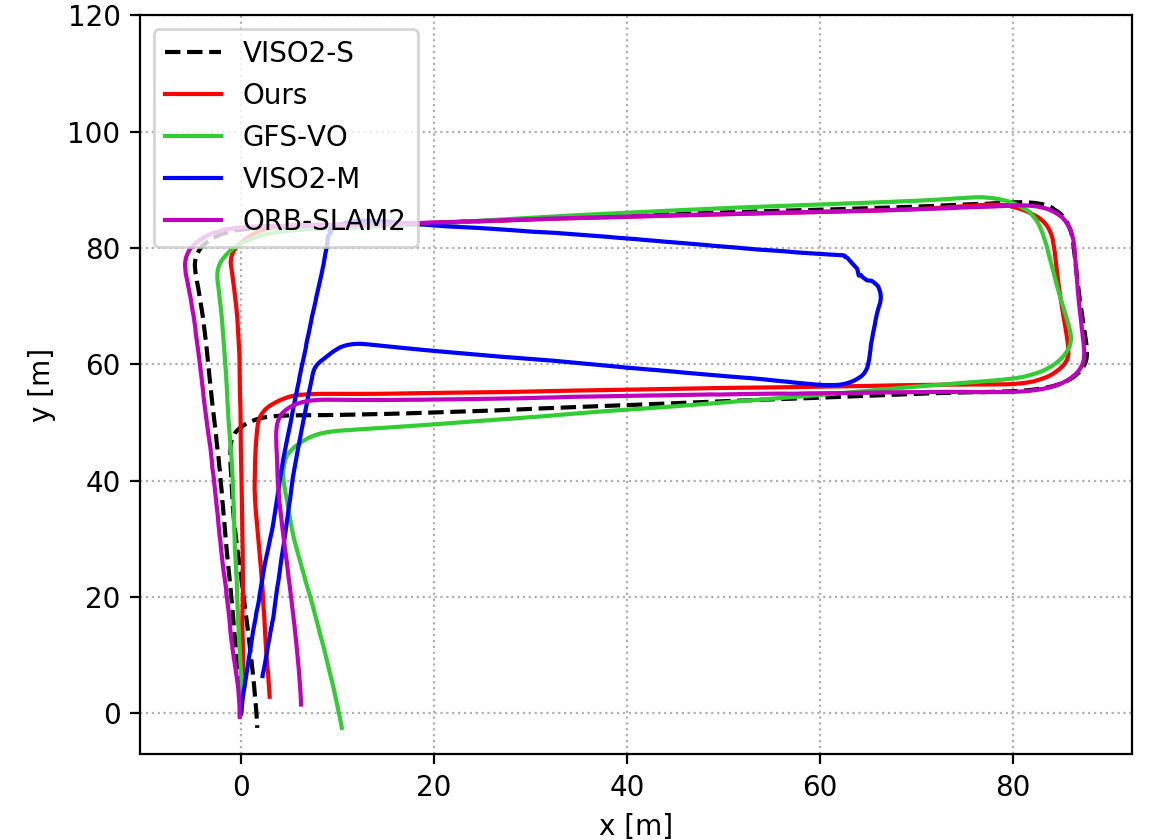}
		\label{fig:trajectory_kitti_14}
	\end{minipage}}
	\subfigure[Seq 15]{
	\begin{minipage}{0.32\textwidth}
		\centering
		\includegraphics[height=4.2cm]{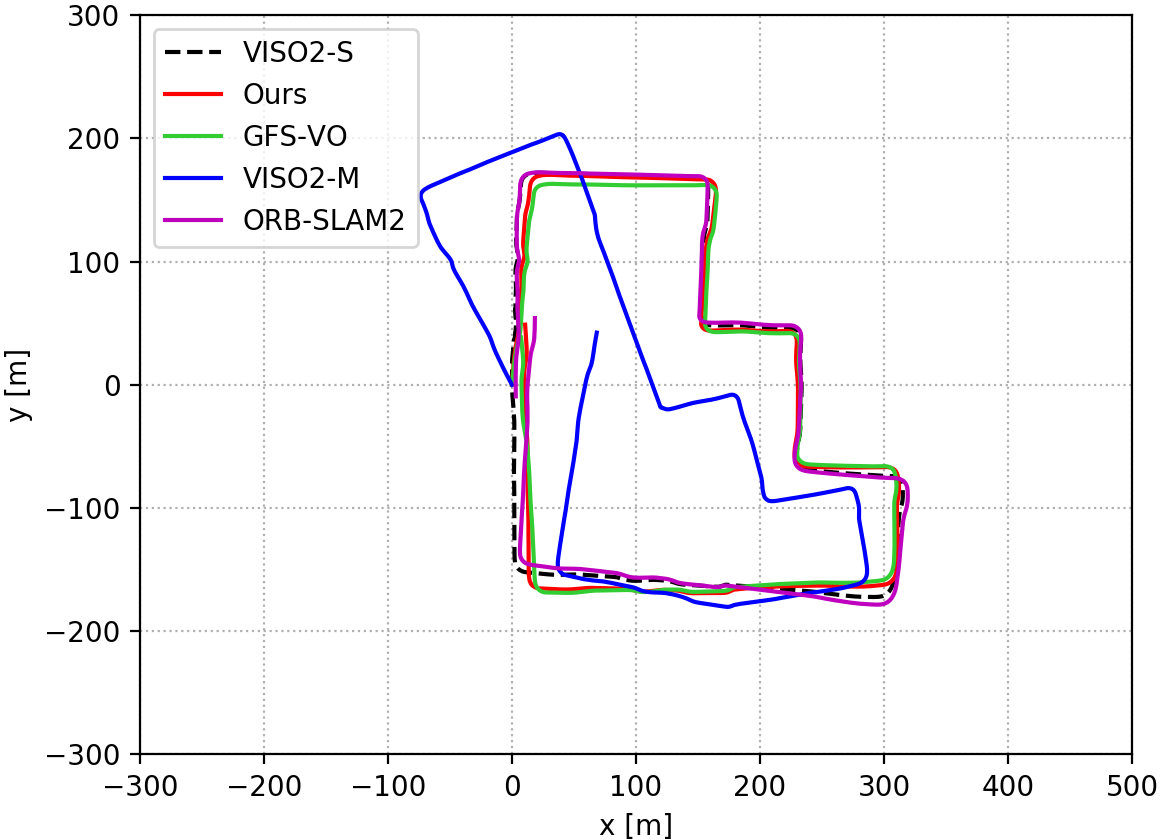}
		\label{fig:trajectory_kitti_15}
	\end{minipage}}
	\subfigure[Seq 16]{
	\begin{minipage}{0.32\textwidth}
		\centering
		\includegraphics[height=4.2cm]{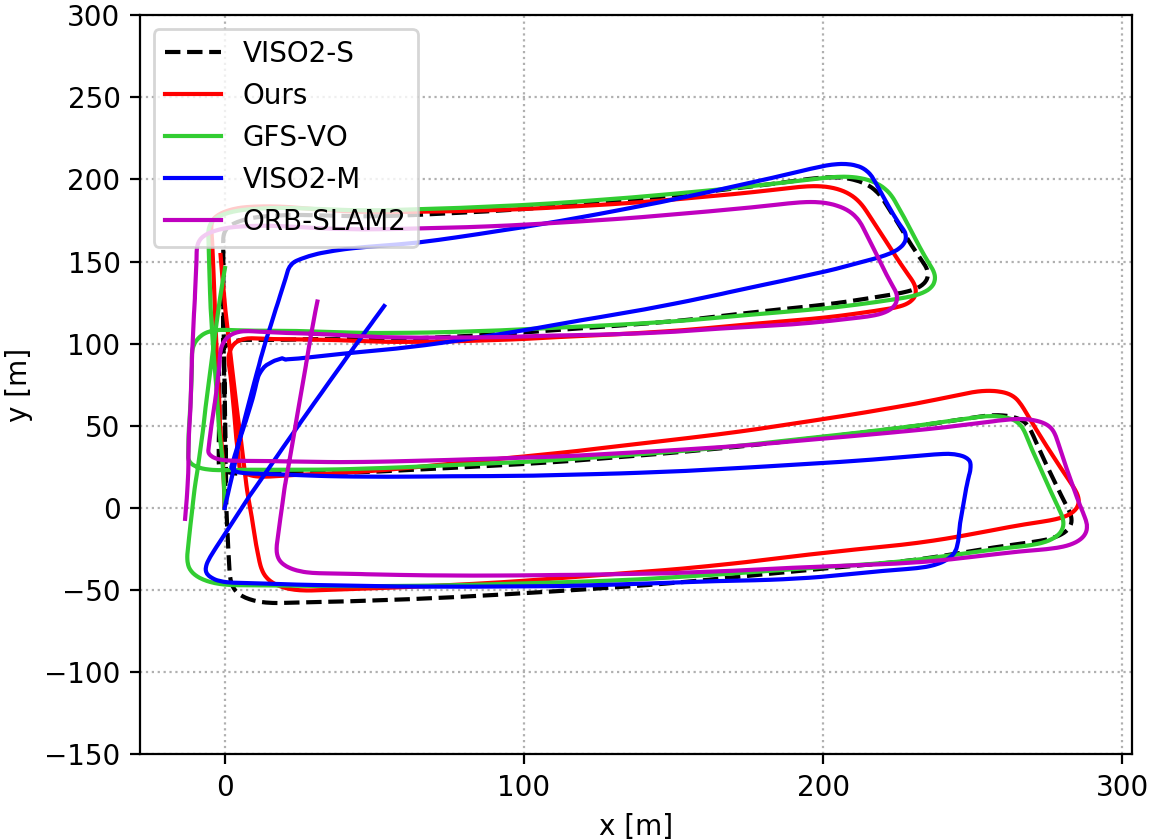}
		\label{fig:trajectory_kitti_16}
	\end{minipage}}
	\subfigure[Seq 17]{
	\begin{minipage}{0.32\textwidth}
		\centering
		\includegraphics[height=4.2cm]{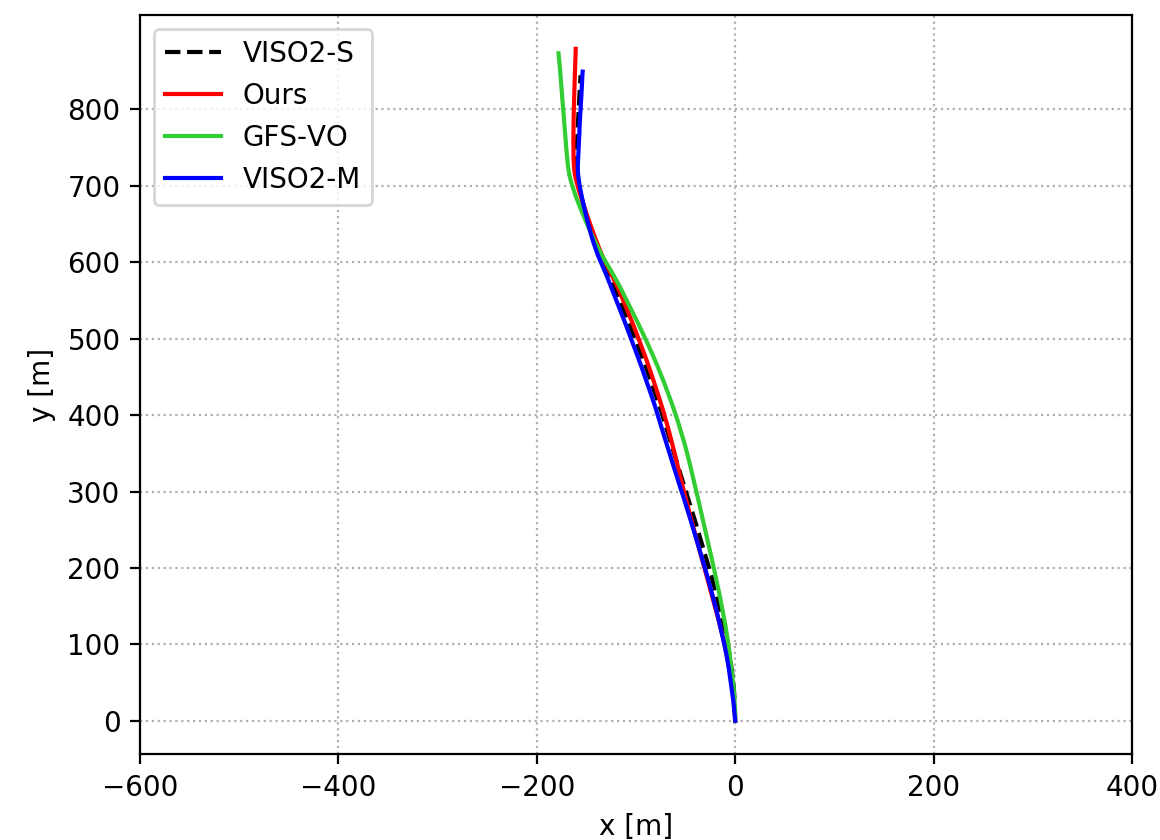}
		\label{fig:trajectory_kitti_17}
	\end{minipage}}
	\subfigure[Seq 18]{
	\begin{minipage}{0.32\textwidth}
		\centering
		\includegraphics[height=4.2cm]{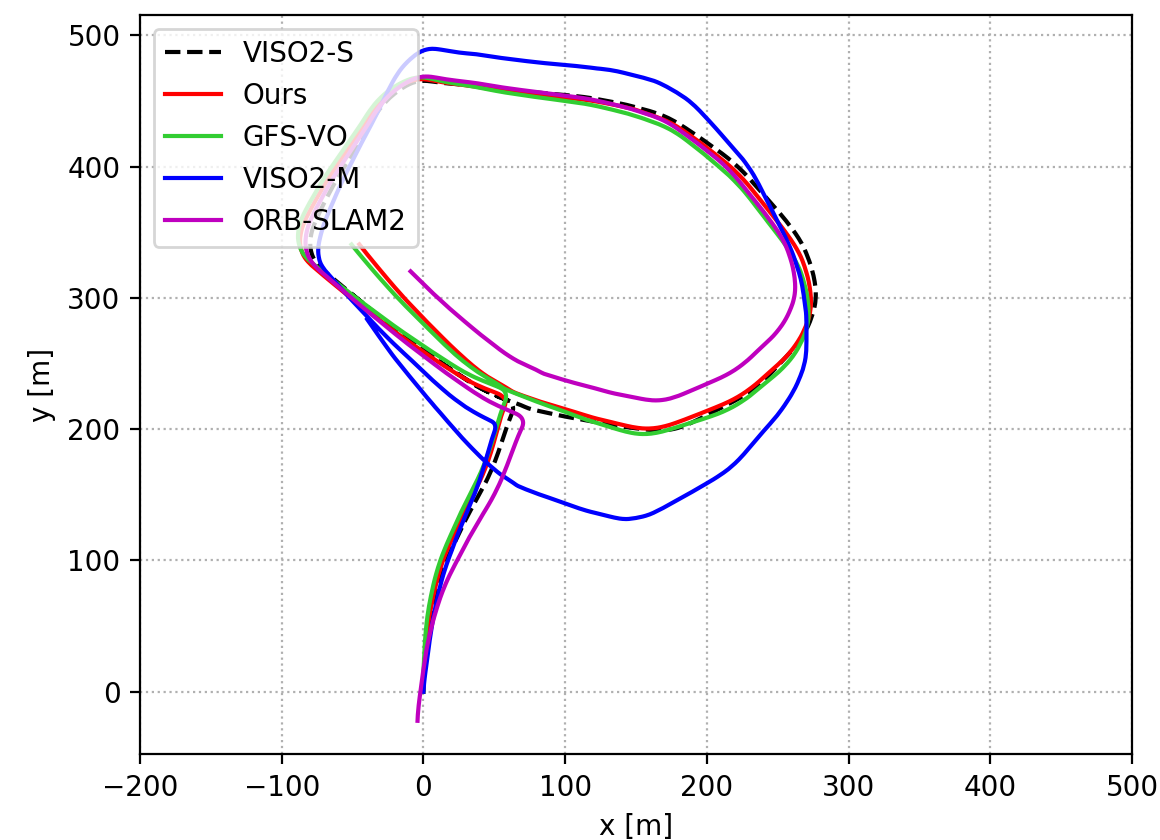}
		\label{fig:trajectory_kitti_18}
	\end{minipage}}
	\subfigure[Seq 19]{
	\begin{minipage}{0.32\textwidth}
	\centering
	\includegraphics[height=4.2cm]{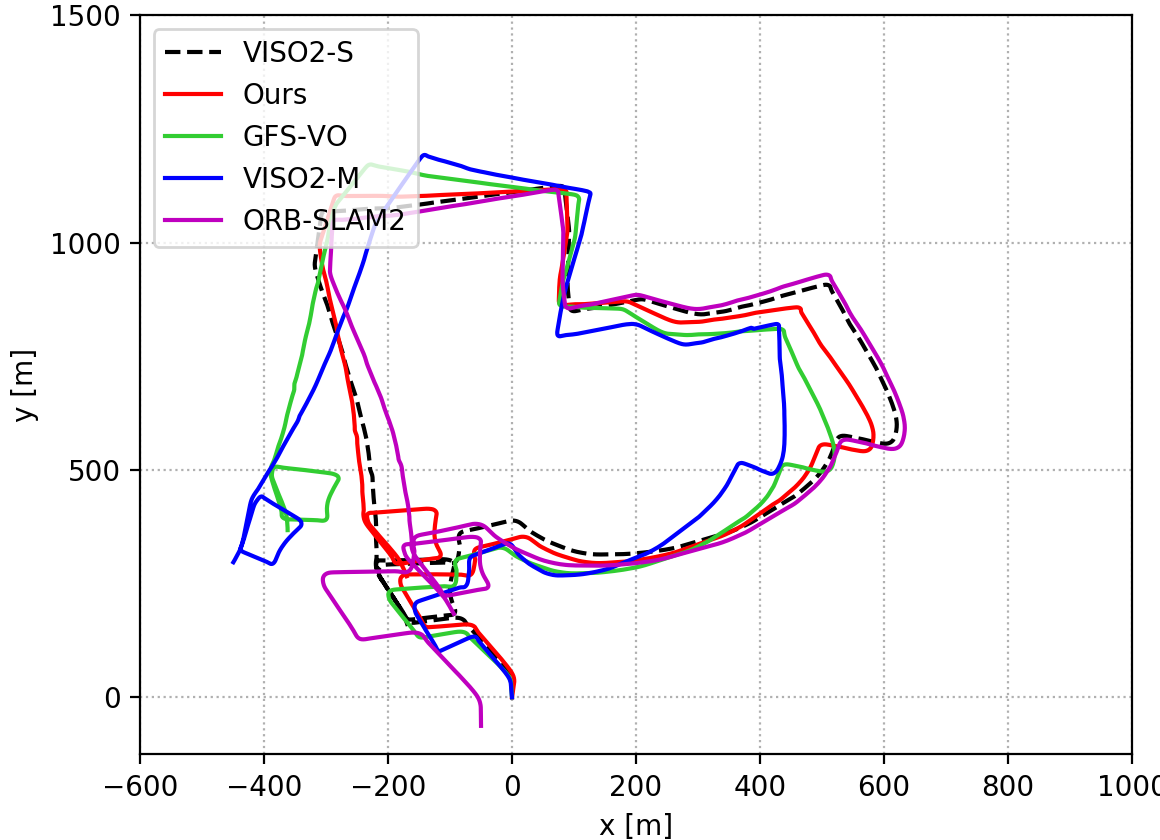}
	\label{fig:trajectory_kitti_19}
	\end{minipage}}
	\caption{The trajectories of stereo VISO2 (VISO2-S) \cite{geiger2011stereoscan}, monocular VISO2 (VISO2-M) \cite{geiger2011stereoscan}, monocular ORB-SLAM2 \cite{mur2017orb-slam2} without loop closure , GFS-VO \cite{xue2018fea}, and our model on Seq 11-19 of the KITTI benchmark dataset \cite{geiger2012kitti}. This time, GFS-VO and our model are trained on the Seq 00-10. As the ground-truths of Seq 11-19 are available, we the results of VISO2-S for references, which is similar to GFS-VO. ORB-SLAM2 fails to initialization in Seq 12 and 17 due to high speeds in highway environments.}
	\label{fig:trajectory_kitti_extra}
\end{figure*}

Qualitative comparison is illustrated in Fig.~\ref{fig:trajectory_kitti_extra}. VISO2-M \cite{geiger2011stereoscan} suffers from severe error accumulation, because it is a classic tracking VO by integrating frame-wise poses as the entire trajectories. ORB-SLAM2 \cite{mur2017orb-slam2} partially alleviates the problem with a global map to assist tracking. Although achieving promising performance in regular environments (Seq 11, 15), it bears large scale drift in complicated scenes (Seq 13, 14, 16, 18, 19). The requirement of sophisticate map initialization degrades its ability to handle situations such as high speeds (Seq 12, 17). 

In contrast, owing to the introduced the \textit{Memory} component for global information gathering and the \textit{Refining} component for ameliorating previous outputs, the scale drift is significantly alleviated in contrast to GFS-VO \cite{xue2018fea}. 


\end{document}